\documentclass[review]{elsarticle}

\usepackage{lineno,hyperref}
\usepackage{stmaryrd}
\usepackage{bm}
\usepackage{booktabs}
\usepackage{graphicx}
\usepackage{float}
\usepackage{graphicx}
\usepackage{stfloats}
\usepackage{amsmath}
\usepackage{epstopdf}
\usepackage{multirow}
\modulolinenumbers[5]

\journal{Journal of \LaTeX\ Templates}

\begin{document}

\begin{frontmatter}

\title{Non-Convex Weighted $\ell_p$ Nuclear Norm based ADMM Framework for Image Restoration}
\tnotetext[mytitlenote]{Fully documented templates are available in the elsarticle package on \href{http://www.ctan.org/tex-archive/macros/latex/contrib/elsarticle}{CTAN}.}

\author{Zhiyuan~Zha$^{a}$, Xinggan~Zhang$^{a}$, Yu~Wu$^{a}$, Qiong~Wang$^{a}$, Lan~Tang$^{a,b}$}
\address{$^{a}$ \footnotesize{School of Electronic Science and Engineering, Nanjing University, Nanjing 210023, China.} \\
    $^{b}$ \footnotesize{National Mobile Commun. Research Lab., Southeast University, Nanjing 210023, China.}}

\begin{abstract}
Since the matrix formed by nonlocal similar patches in a natural image is of low rank, the nuclear norm minimization (NNM) has been widely used in various image processing studies. Nonetheless, nuclear norm based convex surrogate of the rank function usually over-shrinks the rank components and makes different components equally, and thus may produce a result far from the optimum. To alleviate the above-mentioned limitations of the nuclear norm, in this paper we propose a new method for image restoration via the non-convex weighted $\ell_p$ nuclear norm minimization (NCW-NNM), which is able to more accurately enforce the image structural sparsity and self-similarity simultaneously. To make the proposed model tractable and robust, the alternative direction multiplier method (ADMM) is adopted to solve the associated non-convex minimization problem. Experimental results on various types of image restoration problems, including image deblurring, image inpainting and image compressive sensing (CS) recovery, demonstrate that the proposed method outperforms many current state-of-the-art methods in both the objective and the perceptual qualities.
\end{abstract}


\end{frontmatter}

\linenumbers

\section{Introduction}

Image restoration (IR) aims to reconstruct a high quality image $\textbf{\emph{X}}$ from its degraded observation $\textbf{\emph{Y}}$, which can be generally expressed as
\begin{equation}
\textbf{\emph{Y}}= \textbf{\emph{H}}\textbf{\emph{X}}+ \boldsymbol\eta
\label{eq:1}
\end{equation} 
where $\textbf{\emph{H}}$ is a non-invertible linear degradation operator and $\boldsymbol\eta$ is the vector of some independent Gaussian white noise. With different settings of matrix $\textbf{\emph{H}}$, various IR problems can be derived from Eq.~\eqref{eq:1}, such as image denoising \cite{1,2,3,4,70} when $\textbf{\emph{H}}$ is an identity matrix, image deblurring \cite{5,6,7,8} when $\textbf{\emph{H}}$ is a blur operator, image inpainting \cite{9,10,11,12} when $\textbf{\emph{H}}$ is a mask and image compressive sensing (CS) recovery when $\textbf{\emph{H}}$ is a random projection matrix \cite{13,14,15,16}. In this work, we focus on the latter three problems.

IR is a typical ill-posed problem. To deal with this issue, image prior knowledge is usually exploited for regularizing the solution to the following minimization,
\begin{equation}
\hat{\textbf{\emph{X}}}=\arg\min_{\textbf{\emph{X}}} \frac{1}{2} ||\textbf{\emph{Y}}-\textbf{\emph{H}}\textbf{\emph{X}}||_2^2 + \lambda \textbf{\emph{R}}(\textbf{\emph{X}})
\label{eq:2}
\end{equation} 
where the first term  above is the data fidelity term and the second term depends on the employed image priors, and $\lambda$ is the regularization parameter. Due to the ill-posed nature of IR, the image prior knowledge plays a critical role in enhancing the performance of IR algorithms. In other words, how to design an effective regularization model to represent the image priors is vital for IR tasks.

The classical regularization models, such as Tikhonov regularization \cite{17} and total variation (TV) regularization \cite{18,19}, exploited the image local structure and high effectiveness to preserve image edges. Nonetheless,  they tended to over-smooth the image and some image details are usually lost.

As an emerging machine learning technique, sparse representation based modeling has been proved to be a promising model for image restoration \cite{20,21,22,23}. It assumes that image/image patch can be precisely represented as a sparse linear combination of basic elements. These elements, called atoms, compose a dictionary \cite{20,21,24,25}. The dictionary is usually learned from a natural image dataset \cite{20,21}. The well known dictionary learning (DL) based methods, such as KSVD \cite{20,21}, ODL \cite{24} and tasked driven DL \cite{25}, have been proposed and applied to image restoration and other image processing tasks.

Image patches that have similar pattern can be spatially far from each other and thus can be collected in the whole image. This so-called nonlocal self-similarity (NSS) prior is the most outstanding priors for image restoration. The seminal work of nonlocal means (NLM) \cite{1} exploited the NSS prior to perform a series of the weighted filtering for image denoising. Due to its effectiveness, a large amount of related developments have been proposed \cite{3,7,16,26,27,28,29}. For instance, BM3D \cite{28} exploited nonlocal similar 2D image patches and 3D transform domain collaborative filtering. Marial $\emph{et al}.$  \cite{3} considered the idea of NSS by simultaneous sparse coding (SSC).  Dong $\emph{et al}.$ \cite{26} proposed the nonlocally centralized sparse representation (NCSR) model for image restoration, which obtained the estimation of the sparse coding coefficients of the original image by the principle of NLM \cite{1}, and then according to those estimates, NCSR, centralized the sparse coding coefficients of the observed image to improve the restoration performance. Zhang $\emph{et al}.$ \cite{29} proposed a group-based sparse representation framework for image restoration.

Recently, image priors based on NSS \cite{1,4,6,7,22,26,27,28,29,30} and low-rank matrix approximation (LRMA) \cite{43,31,32,39,33,34} have achieved a great success in IR \cite{35,36,37}. A flurry of IR have been proposed, such as image alignment \cite{35}, image/video denoising \cite{4,36,38}, image deblurring \cite{6,40,41} and image inpainting \cite{9,42}. However, these methods usually suffer from a common drawback that the nuclear norm is usually adopted as a convex surrogate of the rank. Despite a good theoretical guarantee by the singular value thresholding (SVT) model \cite{43}, the nuclear norm minimization (NNM) \cite{43,31,33} tends to over-shrink the rank components and treats the different rank components equally, and thus it cannot  approximate the matrix rank accurately enough. To enforce the low rank regularization efficiently, inspired by the success of $\ell_p$ ($0<p<1$) sparse optimization \cite{44,45,46}, Schatten $p$-norm is proposed \cite{47,48,49}, which is defined as the $\ell_p$-norm ($0<p<1$) of the singular values. Compared with traditional nuclear norm, Schatten $p$-norm not only achieves a more accurate recovery result of the signal, but also requires only a \emph{weaker restricted isometry property} based on theory \cite{48}. Nonetheless,  similar to the standard nuclear norm, most of the Schatten $p$-norm based models treat all singular values equally, which may be infeasible in executing many practical problems, such as image inverse problems \cite{50}. To further improve the flexibility of NNM,  Gu $\emph{et al}.$ \cite{4} proposed the weighted nuclear norm minimization (WNNM) model.  Actually, the weighted nuclear norm is essentially the reweighted $\ell_1$-norm of the singular values. Compared with NNM, WNNM assigns  different weights to different singular values such that the matrix rank approximation become more reasonable.

Inspired by the success of $\ell_p$ ($0<p<1$) \cite{44,45,46} and the reweighted $\ell_1$ sparse optimization \cite{51}, to obtain the rank approximation more accurately, we propose a novel model for image restoration (IR) via the non-convex weighted $\ell_p$ ($0<p<1$) nuclear norm minimization (NCW-NNM), which is expected to be more accurate than traditional nuclear norm. Moreover, to solve the associated non-convex minimization problem, we develop an efficient alternative direction multiplier method (ADMM). Experimental results on three typical IR tasks, including image deblurring, image inpainting and image compressive sensing (CS) recovery, show that the proposed method outperforms many current state-of-the-art methods both quantitatively and qualitatively.

The remainder of this paper is organized as follows. Section~\ref{2} introduces  the proposed non-convex weighted $\ell_p$ nuclear norm prior model for image restoration. Section~\ref{3} presents the implementation details of the proposed non-convex model under the ADMM optimization framework. Section~\ref{4} reports the experimental results. Finally, Section~\ref{5} concludes this paper.

\section {Non-convex Weighted $\ell_p$ Nuclear Norm Prior Model for Image Restoration}
\label{2}
Generally speaking, the low rank property of the data matrix formed by nonlocal similar patches in image restoration (IR), is usually characterized by the nuclear norm. However, nuclear norm based convex surrogate of the rank function usually over-shrinks the rank components and makes different components equally, and thus may produce a result far from the optimum. To boost the accuracy of the rank approximation in IR, we propose an efficient IR method via the non-convex weighted $\ell_p$ nuclear norm minimization (NCW-NNM).

In this section, we will elaborate the non-convex weighted $\ell_p$ nuclear norm prior model for IR. Specifically, the basic idea is that data matrix composed of nonlocal similar patches in a natural image is of low rank. The well-known nonlocal self-similarity (NSS) \cite{1,4,6,7,22,26,27,28,29,30}, which depicts the repetitiveness of textures and structures reflected by a natural image within nonlocal regions, implies that many similar patches can be found for any exemplar patch. More specifically, image  $\textbf{\emph{X}}$ with size $\emph{N}$ is divided into $\emph{n}$ overlapped patches $\textbf{\emph{x}}_i$ of size $\sqrt{d}\times\sqrt{d}, i=1,2,...,n$.  Then for each exemplar patch $\textbf{\emph{x}}_i$, its most similar $m$ patches are selected from an $L \times L$ sized searching window to form a set ${\textbf{\emph{S}}}_i$. After this, all the patches in ${\textbf{\emph{S}}}_i$ are stacked into a data matrix ${\textbf{\emph{X}}}_i\in\Re^{{d}\times {m}}$, which contains every element of ${\textbf{\emph{S}}}_i$ as its column, i.e., ${\textbf{\emph{X}}}_i=\{{\textbf{\emph{x}}}_{i,1}, {\textbf{\emph{x}}}_{i,2}, ..., {\textbf{\emph{x}}}_{i,m}\}$, where ${\textbf{\emph{x}}_{i,m}}$ denotes the $m$-th similar patch (column vector form) of the $i$-th group.  Since all the patches have the similar structures in each data matrix, thus, the constructed data matrix ${\textbf{\emph{X}}}_i$ has a low-rank property. Therefore, by incorporating the low-rank prior into Eq.~\eqref{eq:2}, IR is turned into solving the following minimization problem,
\begin{equation}
\hat{\textbf{\emph{X}}}=\arg\min_{\textbf{\emph{X}}} \frac{1}{2} ||\textbf{\emph{Y}}-\textbf{\emph{H}}\textbf{\emph{X}}||_2^2 + \lambda \sum_{i=1}^n \textbf{{Rank}}(\textbf{\emph{X}}_i)
\label{eq:3}
\end{equation} 

In general, the rank minimization is an NP-hard problem. Most of methods resort to using the nuclear norm minimization (NNM) \cite{43,31,33} as a convex relaxation of the non-convex rank minimization. However, since the singular values have clear meanings and should be treated differently, NNM regularizes each of them equally, which cannot achieve the approximation of the matrix rank accurately.  Inspired by the success of $\ell_p$ ($0<p<1$) \cite{44,45,46} and the reweighted $\ell_1$ sparse optimization \cite{51}, we introduce a more flexible non-convex weighted $\ell_p$ nuclear norm  prior model. To be concrete, the weighted $\ell_p$ nuclear norm of a matrix $\textbf{\emph{X}}_i\in\Re^{d\times m}$, which is defined as
\begin{equation}
\textbf{\emph{F}}(\textbf{\emph{X}}_i)=\left(\sum\nolimits_{j=1}^{min\{d,m\}}w_{i,j}\sigma_{i,j}(\textbf{\emph{X}}_i)^p \right)^{\frac{1}{p}}
\label{eq:4}
\end{equation}
where $0<p< 1$, and $\sigma_{i,j}(\textbf{\emph{X}}_i)$ is the $j$-th singular value of a matrix $\textbf{\emph{X}}_i\in\Re^{d\times m}$.  $w_{i,j}\geq0$ is a non-negative weight assigned to $\sigma_{i,j}(\textbf{\emph{X}}_i)$. Then the weighted $\ell_p$ nuclear norm of $\textbf{\emph{X}}_i$ with power $p$ is
\begin{equation}
\textbf{\emph{F}}(\textbf{\emph{X}}_i)= \sum\nolimits_{j=1}^{min\{d,m\}}w_{i,j}\sigma_{i,j}(\textbf{\emph{X}}_i)^p =Tr(\textbf{\emph{W}}_i{\boldsymbol\Sigma}_i^p)
\label{eq:5}
\end{equation}
where $\textbf{\emph{W}}_i$ and ${\boldsymbol\Sigma}_i$ are diagonal matrices whose diagonal entries are composed of all $w_{i,j}$ and $\sigma_{i,j}(\textbf{\emph{X}}_i)$, respectively.

Therefore, considering all the data matrices $\{\textbf{\emph{X}}_i\}$, the  proposed non-convex low rank  prior model for IR is formulated as
\begin{equation}
\hat{\textbf{\emph{X}}}=\arg\min_{\textbf{\emph{X}}} \frac{1}{2} ||\textbf{\emph{Y}}-\textbf{\emph{H}}\textbf{\emph{X}}||_2^2 + \lambda\sum_{i=1}^n \textbf{\emph{F}}(\textbf{\emph{X}}_i)
\label{eq:6}
\end{equation} 

Obviously, it can be seen that the proposed non-convex low-rank prior model is able to employ the structured sparsity of nonlocal similar patches and the non-convexity of rank minimization simultaneously, which is expected to obtain better rank approximation results than many existing methods.

\section{Non-convex Weighted $\ell_p$ Nuclear Norm based ADMM Framework for Image Restoration}
\label{3}
 In this section, the proposed scheme is used to solve  the IR tasks, including image deblurring, image inpainting and image compressive sensing (CS) recovery.  To be concrete, it can be seen that solving the objective function of Eq.~\eqref{eq:6} is very difficult, since it is a large scale non-convex optimization problem. To make the proposed scheme tractable and robust, in this paper, we adopt the alternating direction method of multipliers (ADMM) \cite{52} to solve Eq.~\eqref{eq:6}.

 The ADMM algorithm is a powerful tool for various large scale optimization problems and its basic idea is to turn the unconstrained minimization problem into a constrained one based on variable splitting. Numerical simulations have shown that it can converge  by only using a small memory footprint, which makes it very attractive for numerous large-scale optimization problems \cite{44,53,54}. We will briefly introduce the ADMM algorithm by considering a constrained optimization,
\begin{equation}
\min\nolimits_{\textbf{\emph{u}}\in\Re^{\bf{N}},  \textbf{\emph{z}}\in\Re^{\bf{M}}}f(\textbf{\emph{u}})+g(\textbf{\emph{z}}),\qquad s.t.\qquad
\textbf{\emph{u}}=\textbf{\emph{z}}
\label{eq:7}
\end{equation} 
where $\textbf{\emph{z}}\in\Re^{\bf{M}\times\bf{N}}$ and $f:{\Re^{\bf{N}}\rightarrow\Re}$, $g:{\Re^{\bf{M}}\rightarrow\Re}$. The complete description of the ADMM is shown in Algorithm 1.

\begin{table}[!htbp]
\centering  
\begin{tabular}{lccc}  
\hline  
\qquad \ \  \textbf{Algorithm 1}: \qquad \ \ ADMM Method.\\
\hline
1.  $\textbf{Initialization}~~~k$, choose\ $\rho>0$, ${\textbf{\emph{u}}}, {\textbf{\emph{z}}}$, and ${\textbf{\emph{c}}}.$\\
2. \ $\textbf{Repeat}$\\
3. \ ${\textbf{\emph{u}}}^{k+1}=\arg\min\limits_{\textbf{\emph{u}}}f(\textbf{\emph{u}})
+\frac{\rho}{2}||\textbf{\emph{u}}-{\textbf{\emph{z}}}^{k}-\textbf{\emph{c}}^{k}||_2^2$;\\
4. \ ${\textbf{\emph{z}}}^{k+1}=\arg\min\limits_{\textbf{\emph{z}}}g({\textbf{\emph{z}}})
+\frac{\rho}{2}||{\textbf{\emph{u}}}^{k+1}-{\textbf{\emph{z}}}-\textbf{\emph{c}}^{k}||_2^2$;\\
5. \ ${\textbf{\emph{c}}}^{k+1}={\textbf{\emph{c}}}^{k}-({\textbf{\emph{u}}}^{k+1}-{\textbf{\emph{z}}}^{k+1})$;\\
6. \ ${k}\leftarrow{k}+1;$\\
7. \ $\textbf{Until}$\qquad stoping\ criterion\ is\ satisfied.\\
\hline
\end{tabular}
\end{table}

Now, by  introducing an auxiliary variable $\textbf{\emph{Z}}$ with the constraint $\textbf{\emph{X}}=\textbf{\emph{Z}}$, Eq.~\eqref{eq:6} can be rewritten as

\begin{equation}
{\textbf{\emph{X}}}^{k+1}=\min\limits_{\textbf{\emph{X}}}\frac{1}{2}||\textbf{\emph{Y}}-\textbf{\emph{H}}{\textbf{\emph{X}}}||_2^2
+\frac{\rho}{2}||\textbf{\emph{X}}-\textbf{\emph{Z}}^{k}-\textbf{\emph{C}}^{k}||_2^2
\label{eq:8}
\end{equation}
\begin{equation}
{\textbf{\emph{Z}}}^{k+1}=\min\limits_{\textbf{\emph{Z}}}\frac{\rho}{2}||{\textbf{\emph{X}}}^{k+1}-{\textbf{\emph{Z}}}-\textbf{\emph{C}}^{k}||_2^2
+\lambda\sum_{i=1}^n \textbf{\emph{F}}(\textbf{\emph{Z}}_i)
\label{eq:9}
\end{equation}
and
\begin{equation}
{\textbf{\emph{C}}}^{k+1}={\textbf{\emph{C}}}^{k}-({\textbf{\emph{X}}}^{k+1}-{\textbf{\emph{Z}}}^{k+1})
\label{eq:10}
\end{equation}\par
It can be seen that the minimization for Eq.~\eqref{eq:6} involves splitting two minimization sub-problems, i.e., $\textbf{\emph{X}}$ and $\textbf{\emph{Z}}$ sub-problems. Next, we will introduce that there is an efficient solution to each sub-problem. To avoid confusion, the subscribe $k$ may be omitted for conciseness.

\subsection{${\textbf{\emph{X}}}$ sub-problem}
Given $\textbf{\emph{Z}}$ , the $\textbf{\emph{X}}$ sub-problem denoted by Eq.~\eqref{eq:8} becomes
\begin{equation}
\min_{\textbf{\emph{X}}}{\textbf{\emph{L}}}_1({{\textbf{\emph{X}}}})= \min\limits_{\textbf{\emph{X}}}\frac{1}{2}||\textbf{\emph{Y}}-\textbf{\emph{H}}{\textbf{\emph{X}}}||_2^2
+\frac{\rho}{2}||\textbf{\emph{X}}-\textbf{\emph{Z}}-\textbf{\emph{C}}||_2^2
\label{eq:11}
\end{equation}

Clearly, Eq.~\eqref{eq:11} has a closed-form solution and its solution can be expressed as
\begin{equation}
\hat{\textbf{\emph{X}}}=({\textbf{\emph{H}}}^{T}{\textbf{\emph{H}}}+\rho{\textbf{\emph{I}}})^{-1}({\textbf{\emph{H}}}^{T}{\textbf{\emph{Y}}}+\rho(\textbf{\emph{Z}}+\textbf{\emph{C}}))
\label{eq:12}
\end{equation} 
where ${\textbf{\emph{I}}}$ represents the identity matrix.

Due to the specific structure of $\textbf{\emph{H}}$ in  image deblurring and image inpainting, Eq.~\eqref{eq:11} can be computed without  matrix inversion efficiently ( more details can be seen in \cite{55}).

However, $\textbf{\emph{H}}$  is a random projection matrix without a special structure in image CS recovery, computing the inverse by Eq.~\eqref{eq:11} at each iteration is too costly to implement numerically. Thus, to avoid computing the matrix inversion, an iterative algorithm is highly desired for solving Eq.~\eqref{eq:11}. In this work, we adopt the gradient descent method \cite{56} with an optimal step to solve Eq.~\eqref{eq:11}, i.e.,
\begin{equation}
\hat{\textbf{\emph{X}}}={\textbf{\emph{X}}}-\mu{\textbf{\emph{q}}}
\label{eq:13}
\end{equation} 
where ${\textbf{\emph{q}}}$ is the gradient direction of the objective function ${\textbf{\emph{L}}}_1({{\textbf{\emph{X}}}})$, and $\mu$ is the optimal step.

Therefore, in image CS recovery, it only requires an iterative calculation of the following equation to solve the ${\textbf{\emph{X}}}$ sub-problem,
\begin{equation}
\hat{\textbf{\emph{X}}}={\textbf{\emph{X}}}-\mu({{\textbf{\emph{H}}}}^{T}{{\textbf{\emph{H}}}}{\textbf{\emph{X}}}
-{{\textbf{\emph{H}}}}^{T}{\textbf{\emph{Y}}}+\rho(\textbf{\emph{X}}-{{\textbf{\emph{Z}}}}-\textbf{\emph{C}}))
\label{eq:14}
\end{equation}
where ${{\textbf{\emph{H}}}}^{T}{{\textbf{\emph{H}}}}$ and ${{\textbf{\emph{H}}}}^{T}{\textbf{\emph{Y}}}$ can be calculated in advance.

\subsection{${\textbf{\emph{Z}}}$ sub-problem}
Given ${\textbf{\emph{X}}}$, similarly, according to Eq.~\eqref{eq:9}, the ${\textbf{\emph{Z}}}$ sub-problem can be rewritten as
\begin{equation}
\min_{{\textbf{\emph{Z}}}}{\textbf{\emph{L}}}_2({\textbf{\emph{Z}}})=
\min\limits_{{\textbf{\emph{Z}}}}
\frac{1}{2}||{\textbf{\emph{Z}}}-\textbf{\emph{R}}||_2^2
+\frac{\lambda}{\rho}\sum_{i=1}^n \textbf{\emph{F}}(\textbf{\emph{Z}}_i)
\label{eq:15}
\end{equation}
where  $\textbf{\emph{R}}=\textbf{\emph{X}}-\textbf{\emph{C}}$. However, due to the high complex non-convex structure of $\textbf{\emph{F}}(\textbf{\emph{Z}}_i)$, it is difficult to solve Eq.~\eqref{eq:15}. To obtain a tractable solution of Eq.~\eqref{eq:15}, in this work, a general assumption is made, with which even a closed-form solution can be obtained. Specifically, $\textbf{\emph{R}}$ can be regarded as some type of noisy observation of $\textbf{\emph{X}}$, and then the assumption is made that each element of
$\textbf{\emph{E}}=\textbf{\emph{Z}}-\textbf{\emph{R}}$ follows an independent zero-mean distribution with variance ${
\delta}^{2}$.  The following conclusion can be proved with this assumption.

\noindent$\textbf{Theorem 1}$ \ \ Define $\textbf{\emph{Z}},\textbf{\emph{R}}\in\Re^{N}$, ${\textbf{\emph{Z}}}_i$, ${{\textbf{\emph{R}}}_i}\in\Re^{d\times m}$, and ${\textbf{\emph{e}}}{(j)}$ as each element of  error vector ${\textbf{\emph{e}}}$, where $\textbf{\emph{e}}=\textbf{\emph{Z}}-\textbf{\emph{R}},  j=1,...,N$. Assume that ${\textbf{\emph{e}}}{(j)}$ follows an independent zero mean distribution with variance ${
\delta}^{2}$, and thus for any $\varepsilon>0$, we can represent the relationship between $\frac{1}{N}||\textbf{\emph{Z}}-\textbf{\emph{R}}||_2^2$ and ${\frac{1}{K}}\sum_{i=1}^n||{\textbf{\emph{Z}}}_i-{\textbf{\emph{R}}}_i||_2^2$  by the following property,
\begin{equation}
\lim_{{N\rightarrow\infty}\atop{K\rightarrow\infty}}{\textbf{\emph{P}}}{\{|\frac{1}{N}||\textbf{\emph{Z}}-\textbf{\emph{R}}||_2^2
-{\frac{1}{K}}\sum\nolimits_{i=1}^n||{\textbf{\emph{Z}}}_i-{\textbf{\emph{R}}}_i||_F^2|<\varepsilon\}}=1
\label{eq:16}
\end{equation}
where ${\textbf{\emph{P}}}(\bullet)$  represents the probability and ${\emph{K}}=\emph{d}\times\emph{m}\times\emph{n}$.

\noindent\emph{Proof:}\ \ Owing to the assumption that ${\textbf{\emph{e}}}{(\textbf{\emph{j}})}$  follows an independent zero mean distribution with variance  $\delta^2$, i.e, $\textbf{\emph{E}}[{\textbf{\emph{e}}}{(\textbf{\emph{j}})}]=0$ and $\textbf{\emph{Var}}[{\textbf{\emph{e}}}{(\textbf{\emph{j}})}]=\delta^2$. Thus, it can be deduced that each ${\textbf{\emph{e}}}{(\textbf{\emph{j}})}^2$  is also independent, and the meaning of each  ${\textbf{\emph{e}}}{(\textbf{\emph{j}})}^2$  is:
\begin{equation}
\textbf{\emph{E}}[{\textbf{\emph{e}}}{(\textbf{\emph{j}})}^2]=\textbf{\emph{Var}}[{\textbf{\emph{e}}}{(\textbf{\emph{j}})}]
+[\textbf{\emph{E}}[{\textbf{\emph{e}}}{(\textbf{\emph{j}})}]]^2=\delta^2, \emph{j}=1,2,...,\emph{N}
\label{eq:17}
\end{equation}\par

By invoking the $\emph{law of Large numbers}$ in probability theory, for any $\varepsilon>0$, it leads to
$\lim\limits_{\emph{N}\rightarrow\infty}{\textbf{\emph{P}}}\{|\frac{1}{\emph{N}}\Sigma_{\emph{j}
=1}^{\emph{N}}{\textbf{\emph{e}}}{(\textbf{\emph{j}})}^2-\delta^2|<\frac{\varepsilon}{2}\}=1$, i.e.,
\begin{equation}
\lim\limits_{\emph{N}\rightarrow\infty}{\textbf{\emph{P}}}\{|\frac{1}{\emph{N}}
||\textbf{\emph{Z}}-\textbf{\emph{R}}||_2^2-\delta^2|<\frac{\varepsilon}{2}\}=1
\label{eq:18}
\end{equation}\par
Next, we denote the concatenation of all the groups ${\textbf{\emph{Z}}}_i$  and ${\textbf{\emph{R}}}_i,\ \emph{i}=1,2,...,\emph{n}$, by ${\textbf{\emph{Z}}}$ and ${\textbf{\emph{R}}}$, respectively. Meanwhile, we denote the error of each element of  ${\textbf{\emph{Z}}}-{\textbf{\emph{R}}}$ by ${\textbf{\emph{e}}}{\emph{(k)}},\ \emph{k}=1,2,...,\emph{K}$. We have also denote ${\textbf{\emph{e}}}{\emph{(k)}}$  following an independent zero mean distribution with variance $\delta^2$.\par
Therefore, the same process applied to ${\textbf{\emph{e}}}{\emph{(k)}}^2$ yields $\lim\limits_{\emph{N}\rightarrow\infty}{\textbf{\emph{P}}}\{|\frac{1}{\emph{N}}\Sigma_{\emph{k}
=1}^{\emph{N}}{\textbf{\emph{e}}}{({\emph{k}})}^2-\delta^2|<\frac{\varepsilon}{2}\}=1$, i.e.,

\begin{equation}
\lim\limits_{\emph{N}\rightarrow\infty}{\textbf{\emph{P}}}\{|\frac{1}{\emph{N}}\Sigma_{\emph{i}
=1}^{\emph{n}}||{\textbf{\emph{Z}}}_{i}-{\textbf{\emph{R}}}_{i}||_2^2-\delta^2|<\frac{\varepsilon}{2}\}=1
\label{eq:19}
\end{equation}\par
Obviously, considering Eqs.~\eqref{eq:18} and \eqref{eq:19} together, we can prove Eq.~\eqref{eq:16}.

Accordingly, based on $\emph{Theorem 1}$, we have the following equation with a very large probability (restricted 1) at each iteration,
\begin{equation}
\frac{1}{N}||\textbf{\emph{Z}}-\textbf{\emph{R}}||_2^2
={\frac{1}{K}}\sum\nolimits_{i=1}^n||{\textbf{\emph{Z}}}_i-{\textbf{\emph{R}}}_i||_F^2
\label{eq:20}
\end{equation}

Therefore, based on Eqs.~\eqref{eq:15} and~\eqref{eq:20}, we have
\begin{equation}
\begin{aligned}
&\min\limits_{{\textbf{\emph{Z}}}}
\frac{1}{2}||{\textbf{\emph{Z}}}-\textbf{\emph{R}}||_2^2
+\frac{\lambda}{\rho}\sum_{i=1}^n \textbf{\emph{F}}(\textbf{\emph{Z}}_i)\\
&=\min\limits_{{\textbf{\emph{Z}}}_i}\sum\nolimits_{i=1}^n \left(\frac{1}{2}||{\textbf{\emph{Z}}}_i-{\textbf{\emph{R}}}_i||_F^2
 +{{\tau}} \textbf{\emph{F}}(\textbf{\emph{Z}}_i) \right)\\
\end{aligned}
\label{eq:21}
\end{equation}
where $\textbf{\emph{F}}(\textbf{\emph{Z}}_i)=\sum\nolimits_{j=1}^{min\{d,m\}}w_{i,j}\sigma_{i,j}(\textbf{\emph{Z}}_i)^p$ and $0<p<1$.  $\sigma_{i,j}(\textbf{\emph{Z}}_i)$ denotes the $j$-th singular value of the matrix $\textbf{\emph{Z}}_i$ and ${{\tau}}={{{\lambda}}{\emph{K}}}/{\rho{\emph{N}}}$. Obviously, one can observe that Eq.~\eqref{eq:21} can be efficiently minimized by solving $n$ sub-problems for all the data matrices $\textbf{\emph{Z}}_i$. However, due to the fact that Eq.~\eqref{eq:21} is high non-convex, it seems to be very difficult to solve it.  Nonetheless, to achieve an effective solution of Eq.~\eqref{eq:21}, we have the following Theorems.

\noindent$\textbf{Theorem 2}$ (von Neumann)\ \ For any two matrices $\textbf{\emph{A}}, \textbf{\emph{B}} \in\Re^{m \times n}$, $tr({\textbf{\emph{A}}}^T\textbf{\emph{B}}) \leq tr(\sigma({\textbf{\emph{A}}})^T\sigma({\textbf{\emph{B}}}))$, where $\sigma({\textbf{\emph{A}}})$ and $\sigma({\textbf{\emph{B}}})$ are the ordered singular value matrices of $\textbf{\emph{A}}$ and $\textbf{\emph{B}}$ with the same order, respectively.

The proof can be seen in \cite{71}.

\noindent$\textbf{Theorem 3}$ \ \ Let $\textbf{\emph{R}}_i= \textbf{\emph{U}}_i\boldsymbol\Delta_i\textbf{\emph{V}}_i^T$ be the SVD of $\textbf{\emph{R}}_i\in\Re^{d\times m}$ and $\boldsymbol\Delta_i=diag(\gamma_{i,1},...,\gamma_{i,j})$, $j= min(d,m)$. The optimal solution $\textbf{\emph{Z}}_i$ to problem Eq.~\eqref{eq:21} is $\textbf{\emph{U}}_i\boldsymbol\Sigma_i\textbf{\emph{V}}_{i}^T$, where $\boldsymbol\Sigma_{i}=diag(\sigma_{i,1},...,\sigma_{i,j})$. Then the optimal solution of the $j$-th diagonal element $\sigma_{i,j}$ of the diagonal matrix $\boldsymbol\Sigma_{i}$ is solved by the following problem,
\begin{equation}
\begin{aligned}
&\min\limits_{\sigma_{i,j}\geq0}\sum_{j=1}^j
\left(\frac{1}{2}(\sigma_{i,j}-\gamma_{i,j})^2+\tau w_{i,j}\sigma_{i,j}^p \right)\\
\end{aligned}
\label{eq:22}
\end{equation}
where $\sigma_{i,j}$ represents the $j$-th singular value of each data matrix $\textbf{\emph{Z}}_i$.

\noindent\emph{Proof:}\ \ Suppose the SVD of $\textbf{\emph{Z}}_i$ and $\textbf{\emph{R}}_i$ are $\textbf{\emph{Z}}_i= \textbf{\emph{P}}_i\boldsymbol\Sigma_i\textbf{\emph{Q}}_{i}^T$ and $\textbf{\emph{R}}_i= \textbf{\emph{U}}_i\boldsymbol\Delta_i\textbf{\emph{V}}_{i}^T$, respectively, where $\boldsymbol\Sigma_i$ and $\boldsymbol\Delta_i$ are ordered singular value matrices with the same order. Then Eq.~\eqref{eq:21} can be rewritten as
\begin{equation}
\min\limits_{{\textbf{\emph{Z}}}_i}\sum\nolimits_{i=1}^n \left(\frac{1}{2}||\textbf{\emph{P}}_i\boldsymbol\Sigma_i\textbf{\emph{Q}}_{i}^T-\textbf{\emph{U}}_i\boldsymbol\Delta_i\textbf{\emph{V}}_{i}^T||_F^2
 +{{\tau}} tr({\textbf{\emph{W}}}_i\boldsymbol\Sigma_i^p) \right)
\label{eq:22.0}
\end{equation}
where ${\textbf{\emph{W}}}_i=\{{w_{i,1}, w_{i,2},..., w_{i,j}}\}^T$  and $w_{i,j}$ is a non-negative weight assigned to $\boldsymbol\Sigma_i$.

Based on Theorem 2, we have
\begin{equation}
\begin{aligned}
&||\textbf{\emph{P}}_i\boldsymbol\Sigma_i\textbf{\emph{Q}}_{i}^T-\textbf{\emph{U}}_i\boldsymbol\Delta_i\textbf{\emph{V}}_{i}^T||_F^2\\
&= tr(\boldsymbol\Sigma_i {\boldsymbol\Sigma_i}^T)
+tr(\boldsymbol\Delta_i {\boldsymbol\Delta_i}^T)-2tr({\textbf{\emph{Z}}}_i^T {\textbf{\emph{R}}}_i)\\
& \geq tr(\boldsymbol\Sigma_i {\boldsymbol\Sigma_i}^T)
+tr(\boldsymbol\Delta_i {\boldsymbol\Delta_i}^T)-2tr({\boldsymbol\Sigma_i}^T \boldsymbol\Delta_i)\\
&=||\boldsymbol\Sigma_i-\boldsymbol\Delta_i||_F^2\\
\end{aligned}
\label{eq:23}
\end{equation}
where the equality holds only when $\textbf{\emph{P}}_{i}= \textbf{\emph{U}}_{i}$ and $\textbf{\emph{Q}}_{i}= \textbf{\emph{V}}_{i}$. Therefore, Eq.~\eqref{eq:21} is minimized when $\textbf{\emph{P}}_{i}= \textbf{\emph{U}}_{i}$ and $\textbf{\emph{Q}}_{i}= \textbf{\emph{V}}_{i}$, and the optimal solution of ${\boldsymbol\Sigma_i}$ is obtained by solving the following problem,
\begin{equation}
\begin{aligned}
& \min_{\boldsymbol\Sigma_i\geq0}\frac{1}{2} ||\boldsymbol\Sigma_i- \boldsymbol\Delta_i||_F^2+ tr({\textbf{\emph{W}}}_i\boldsymbol\Sigma_i^p)\\
&= \min\limits_{\sigma_{i,j}\geq0}\sum_{j=1}^j
\left(\frac{1}{2}(\sigma_{i,j}-\gamma_{i,j})^2+\tau w_{i,j}\sigma_{i,j}^p \right)\\
\end{aligned}
\label{eq:24}
\end{equation}

Therefore, the minimization problem of Eq.~\eqref{eq:21} can be simplified by minimizing the problem of Eq.~\eqref{eq:22}.

Therefore, based on Theorem 2, the problem of Eq.~\eqref{eq:21} is transformed into solving Eq.~\eqref{eq:22}. To obtain the solution of Eq.~\eqref{eq:22} effectively, in this subsection, the generalized soft-thresholding (GST) algorithm \cite{45} is used to solve Eq.~\eqref{eq:22}. More specifically, given $p$, $\gamma_{i,j}$ and $w_{i,j}$, there exists a specific threshold,
\begin{equation}
\tau_p^{\emph{GST}}({{{{w}}}_{i,j}})=(2{{{{w}}}_{i,j}}(1-p))^{\frac{1}{2-p}}+{{{{w}}}_{i,j}}p(2{{{{w}}}_{i,j}}(1-p))^{\frac{p-1}{2-p}}
\label{eq:25}
\end{equation} 

Here if ${{\gamma}_{i,j}}<\tau_p^{\emph{GST}}({{{{w}}}_{i,j}})$, ${{\sigma}_{i,j}}=0$ is the global minimum. Otherwise, the optimum will be obtained at non-zero point. According to \cite{45}, for any ${{\gamma}_{i,j}}\in(\tau_p^{\emph{GST}}({{{{w}}}_{i,j}}), +\infty)$, Eq.~\eqref{eq:22} has one unique minimum ${\textbf{\emph{T}}}_p^{\emph{GST}}({{\gamma}_{i,j}}; {{{{w}}}_{i,j}})$, which can be obtained by solving the following equation,
\begin{equation}
{\textbf{\emph{T}}}_p^{\emph{GST}}({{\gamma}_{i,j}}; {{{{w}}}_{i,j}})- {{\gamma}_{i,j}} + {{{{w}}}_{i,j}}p \left({\textbf{\emph{T}}}_p^{\emph{GST}}({{\gamma}_{i,j}}; {{{{w}}}_{i,j}})\right)^{p-1} =0
\label{eq:26}
\end{equation} 
The complete description of the GST algorithm is exhibited in Algorithm 2.  For more details about the GST algorithm, please refer to \cite{45}.

\begin{table}[!htbp]
\centering  
\begin{tabular}{lccc}  
\hline  
\qquad \ \  \textbf{Algorithm 2}: Generalized Soft-Thresholding (GST) \cite{45}.\\
\hline
$\textbf{Input:}$ \ ${{\gamma}_{i,j}}, {{{{w}}}_{i,j}}, p, J$.\\
1.\ \ \ $\tau_p^{\emph{GST}}({{{{w}}}_{i,j}})=(2{{{{w}}}_{i,j}}(1-p))^{\frac{1}{2-p}}+{{{{w}}}_{i,j}}p(2{{{{w}}}_{i,j}}(1-p))^{\frac{p-1}{2-p}}
$;\\
2.\ \ \ $\textbf{If}$ \ \ \ $|{{\gamma}_{i,j}}|\leq \tau_p^{\emph{GST}}({{{{w}}}_{i,j}})$\\
3.\ \ \ \ \ \ \ \ \ ${\textbf{\emph{T}}}_p^{\emph{GST}}({{\gamma}_{i,j}}; {{{{w}}}_{i,j}})=0$;\\
4.\ \ \ $\textbf{else}$\\
5.\ \ \ \ \ \ \ \ \ $t=0, {{\sigma}_{i,j}}^{(t)}=|{{\gamma}_{i,j}}|$;\\
6.\ \ \ \ \ \ \ \ \ Iterate on $t=0, 1, ..., J$\\
7.\ \ \ \ \ \ \ \ \ ${{\sigma}_{i,j}}^{(t+1)}=|{{\gamma}_{i,j}}|-{{{{w}}}_{i,j}}p \left({{\sigma}_{i,j}}^{(t)}\right)^{p-1}$;\\
8.\ \ \ \ \ \ \ \ \ $t\leftarrow t+1$;\\
9.\ \ \ \ \ \ \ \ \ ${\textbf{\emph{T}}}_p^{\emph{GST}}({{\gamma}_{i,j}}; {{{{w}}}_{i,j}})={\rm sgn}({{\gamma}_{i,j}}){{\sigma}_{i,j}}^{t}$;\\
10.\ \  $\textbf{End}$\\
$\textbf{Input:}$: ${\textbf{\emph{T}}}_p^{\emph{GST}}({{\gamma}_{i,j}}; {{{{w}}}_{i,j}})$.\\
\hline
\end{tabular}
\end{table}

Therefore, a closed-form solution of  Eq.~\eqref{eq:22} can be computed as
 \begin{equation}
{{{{\sigma}}}_{i,j}}={{\emph{GST}}}(\gamma_{i,j}, \tau w_{i,j}, p, J)
\label{eq:27}
\end{equation}
where $J$ denotes the iteration number of the GST algorithm.

For each weight $w_{i,j}$, large singular values of a group $\textbf{\emph{R}}_i$ transmit  major edge and texture information. This implies that to reconstruct ${\textbf{\emph{X}}}_i$ from its degraded one, we should shrink the larger singular values less, while shrinking smaller ones more. Therefore, we let
\begin{equation}
{{{{w}}}}_{i,j}= \frac{1}{|{{{\boldsymbol\gamma}}}_{i,j}|+\epsilon}
\label{eq:28}
\end{equation}
where $\epsilon$ is a small constant.

The parameter $\lambda$ that balances the fidelity term and the regularization term should be adaptively determined for better reconstruction performance. In this paper, inspired by \cite{57}, the regularization parameter $\lambda$ of each group ${\textbf{\emph{R}}}_i$ is set as:
\begin{equation}
\lambda=\frac{2\sqrt{2}\delta^2}{\theta_{i}+\varsigma}
\label{eq:29}
\end{equation}
where $\theta_{i}$ denotes the estimated variance of $\boldsymbol\Delta_i$, and $\varsigma$ is a small constant.

\subsection {Summary of the Algorithm}
 The above two sub-problems $\textbf{\emph{X}}$, $\textbf{\emph{Z}}$ have been solved. We can achieve an efficient solution by solving each sub-problem separately, which can guarantee the whole algorithm to be more efficient and effective. Based on the above analysis, the complete description of the proposed method for the IR via non-convex weighted $\ell_p$ nuclear norm prior model is exhibited in Table~\ref{lab:1}.

 \begin{table}[!htbp]
\caption{The Proposed NCW-NNM model for Image Restoration.}
\centering  
\begin{tabular}{lccc}  
\hline
$\textbf{Input:}$ \ The observed image $\textbf{\emph{Y}}$, the degraded operator \textbf{\emph{H}}.\\
  $\rm \textbf{Initialization:} $\  $k$, \textbf{\emph{C}}, \textbf{\emph{Z}}, \emph{d}, \emph{m}, $\rho$, \emph{p}, $\delta$,  $\epsilon$, $\varsigma$, \emph{L}, \emph{K};\\
  $\rm \textbf{Repeat}$\\
    \qquad $\rm \textbf{If}$ $\textbf{\emph{H}}$ is blur operator or mask operator\\
      \qquad \qquad Update $\textbf{\emph{X}}^{k+1}$ computing by Eq.~\eqref{eq:12};\\
    \qquad $\rm \textbf{Else}$ $\textbf{\emph{H}}$ is random projection operator\\
  \qquad \qquad Update $\textbf{\emph{X}}^{k+1}$ computing by Eq.~\eqref{eq:14};\\
      \qquad $\rm \textbf{End if}$\\
    \qquad \qquad $\textbf{\emph{R}}^{k+1}=\textbf{\emph{X}}^{k+1}-\textbf{\emph{C}}^{k}$;\\
   \qquad \qquad Generating the groups $\{\textbf{\emph{R}}_i\}$ by searching similar patches from $\textbf{\emph{R}}$;\\
 $\rm \textbf{For}$\qquad \ Each group $\textbf{\emph{R}}_i$ $\rm \textbf{do}$\\
 \qquad \qquad Singular value decomposition $[\textbf{\emph{U}}_i, \boldsymbol\Delta_i, \textbf{\emph{V}}_i]=\emph{SVD}(\textbf{\emph{R}}_i)$;\\
 \qquad \qquad Update the weight ${\textbf{\emph{W}}_i}^{k+1}$ computing by Eq.~\eqref{eq:28}\\
  \qquad \qquad Update $\lambda^{k+1}$ computing by Eq.~\eqref{eq:29};\\
   \qquad \qquad Calculate $\boldsymbol\Sigma_i$ by using Eq.~\eqref{eq:27};\\
   \qquad \qquad Get the estimation ${{{\textbf{\emph{Z}}}}_i}^{k+1}$ =$\textbf{\emph{U}}_i \boldsymbol\Sigma_i {\textbf{\emph{V}}_i}^T$;\\
  $\rm \textbf{End for}$\\
   \qquad \qquad Update $\textbf{\emph{Z}}^{k+1}$ by aggregating all groups ${{{\textbf{\emph{Z}}}}_i}^{k+1}$;\\
   \qquad \qquad Update $\textbf{\emph{C}}^{k+1}$ by computing Eq.~\eqref{eq:10};\\
      \qquad \qquad $k\leftarrow k+1$;\\
    $\rm \textbf{End for}$\\
         $\textbf{Until}$\\
       \qquad      \qquad maximum iteration number is reached.\\
     $\textbf{Output:}$\\
    \qquad \qquad    The final restored image ${\hat{\textbf{\emph{X}}}}$.\\
\hline
\end{tabular}
\label{lab:1}
\end{table}

\section {Experimental Results}
\label{4}
\begin{figure*}[!htbp]
\begin{minipage}[b]{1\linewidth}
  \centerline{\includegraphics[width=12cm]{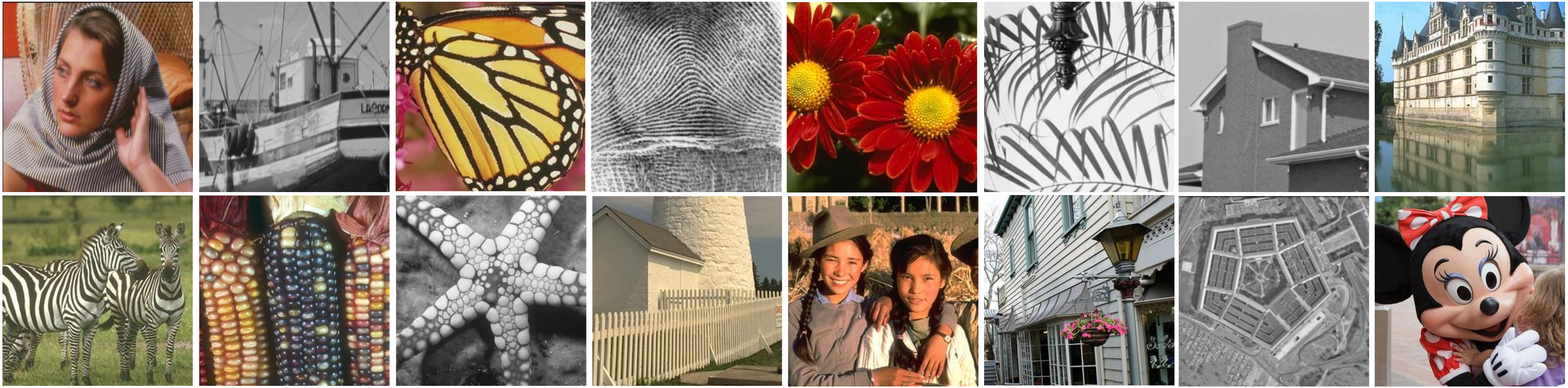}}
\end{minipage}
\caption{All test images. Top: Barbara, boats, Butterfly, fingerprint, Flower, Leaves, House, Castle. Below: Zebra, Corn, starfish, Fence, Girl, Light, pentagon, Mickey.}
\label{fig:1}
\end{figure*}

In this section, we conduct a variety of experiments on three IR tasks, including image deblurring, image inpainting and image CS recovery, to evaluate the effectiveness of  the proposed NCW-NNM method. To prove the advantage of the proposed NCW-NNM method, which can improve the accuracy of the matrix rank approximation. We compare it with  traditional rank minimization method, i.e., nuclear norm minimization (NNM) method. All the experimental images are shown in Fig.~\ref{fig:1}. To evaluate the quality of the restored images, the PSNR and the recently proposed powerful perceptual quality metric FSIM \cite{58} are calculated. All the experiments were implemented in Matlab 2012b on a PC with 3.50GHz CUP and 4GB RAM. 

\subsection{Image Deblurring}

In image deblurring, two blur kernels, $9\times 9$ uniform kernel and a Gaussian blur kernel (fspecial('gaussian', 25, 1.6)) are used. Blurred images are further corrupted by additive white Gaussian noise with $\delta=\sqrt{2}$. The parameters are set as follows. The size of each patch $\sqrt{d} \times \sqrt{d}$ is set to $8\times 8$. Similar patch numbers $m=60$, $L=25$, $\epsilon=0.1$, $\varsigma=0.3$, $J=2$. ($\rho, p$) are set to (0.02, 0.7) and (0.06, 0.6) for Gaussian blur and Uniform blur, respectively.

\begin{figure*}[!htbp]
\begin{minipage}[b]{1\linewidth}
  \centerline{\includegraphics[width=12cm]{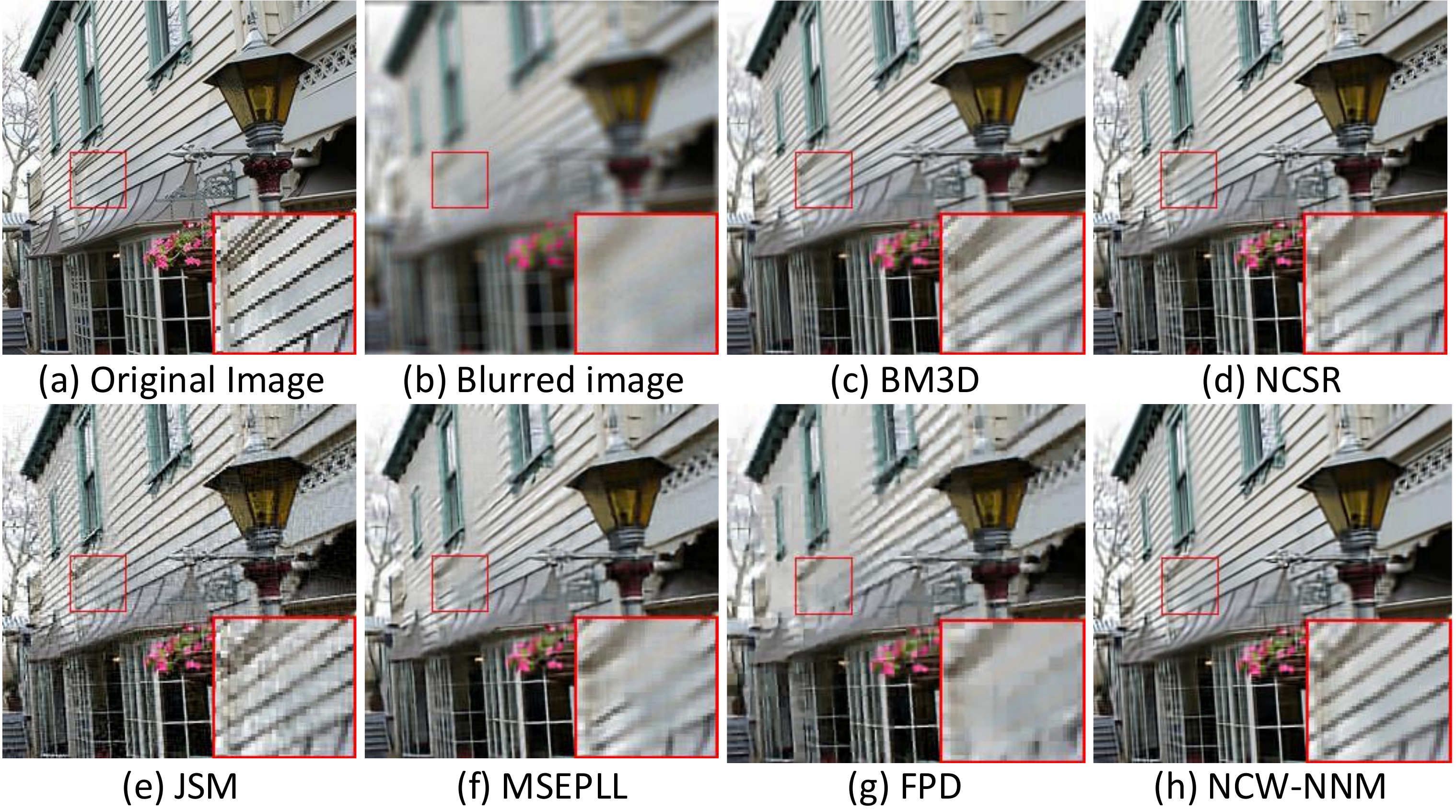}}
\end{minipage}
\caption{Visual comparison  of \emph{Light} by image deblurring with $9\times 9$ Uniform Kernel, $\delta=\sqrt{2}$. (a) Original image; (b) Noisy and blurred image; (c) deblurred image by (c) BM3D \cite{59} (PSNR = 21.65dB, FSIM= 0.8423); (d) NCSR \cite{26} (PSNR = 22.10dB, FSIM = 0.8733); (e) JSM \cite{7} (PSNR =21.84dB, FSIM = 0.8439); (f) MSEPLL \cite{60} (PSNR = 21.27dB, FSIM = 0.8417); (g) FPD \cite{61} (PSNR = 19.70dB, FSIM = 0.7725); (h) NCW-NNM (PSNR =\textbf{22.58dB}, FSIM =\textbf{0.8790}).}
\label{fig:2}
\end{figure*}

\begin{figure*}[!htbp]
\begin{minipage}[b]{1\linewidth}
  \centerline{\includegraphics[width=12cm]{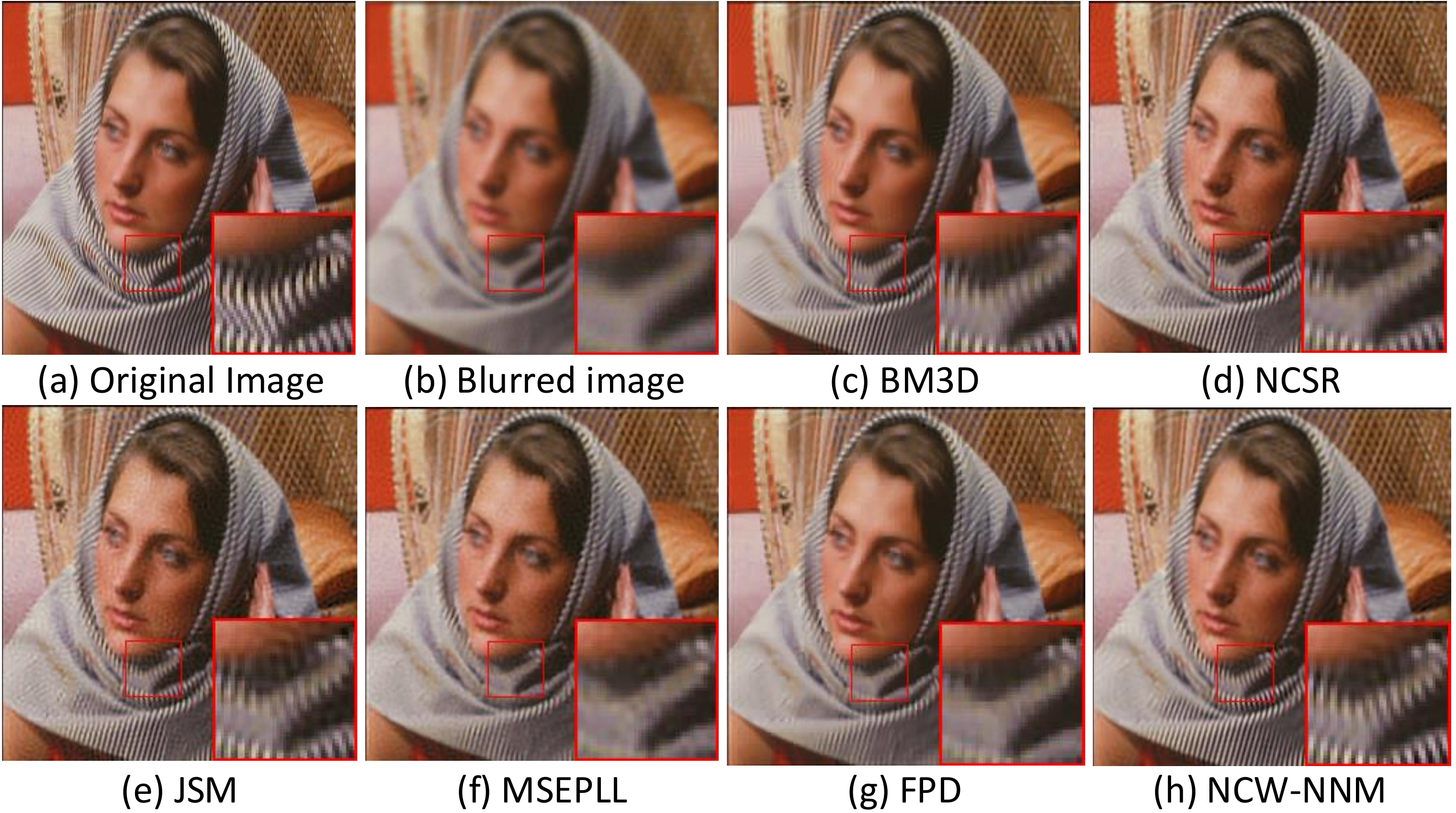}}
\end{minipage}
\caption{Visual comparison  of \emph{Barbara} by image deblurring with Gaussian blur: fspecial('gaussian', 25, 1.6), $\delta=\sqrt{2}$. (a) Original image; (b) Noisy and blurred image; (c) deblurred image by (c) BM3D \cite{59} (PSNR = 25.77dB, FSIM= 0.8802); (d) NCSR \cite{26} (PSNR = 25.95dB, FSIM = 0.8853); (e) JSM \cite{7} (PSNR = 25.89dB, FSIM = 0.8752); (f) MSEPLL \cite{60} (PSNR = 24.82dB, FSIM = 0.8579); (g) FPD \cite{61} (PSNR = 24.54dB, FSIM = 0.8622); (h) NCW-NNM (PSNR =\textbf{26.72dB}, FSIM =\textbf{0.9009}).}
\label{fig:3}
\end{figure*}

We have compared the proposed NCW-NNM with five other competing methods including BM3D \cite{59}, NCSR \cite{26}, JSM \cite{7}, MSEPLL \cite{60} and FPD \cite{61}.  The PSNR and FSIM results are shown in Table~\ref{lab:2}.  The average gain of the proposed NCW-NNM over BM3D, NCSR, JSM, MSEPLL and FPD methods can be as much as 0.64dB, 0.16dB, 1.07dB, 0.79dB and 1.85dB, respectively. The visual comparisons of the deblurring methods are shown in Fig.~\ref{fig:2} and Fig.~\ref{fig:3}. It can be seen that BM3D, NCSR, JSM, MSEPLL and FPD still generate some undesirable artifacts, while result in over-smooth phenomena. By contrast, the proposed NCW-NNM not only preserves the sharpness of edges, but also suppresses undesirable artifacts more efficiently.
\begin{figure*}[!htbp]
\begin{minipage}[b]{1\linewidth}
  \centerline{\includegraphics[width=12cm]{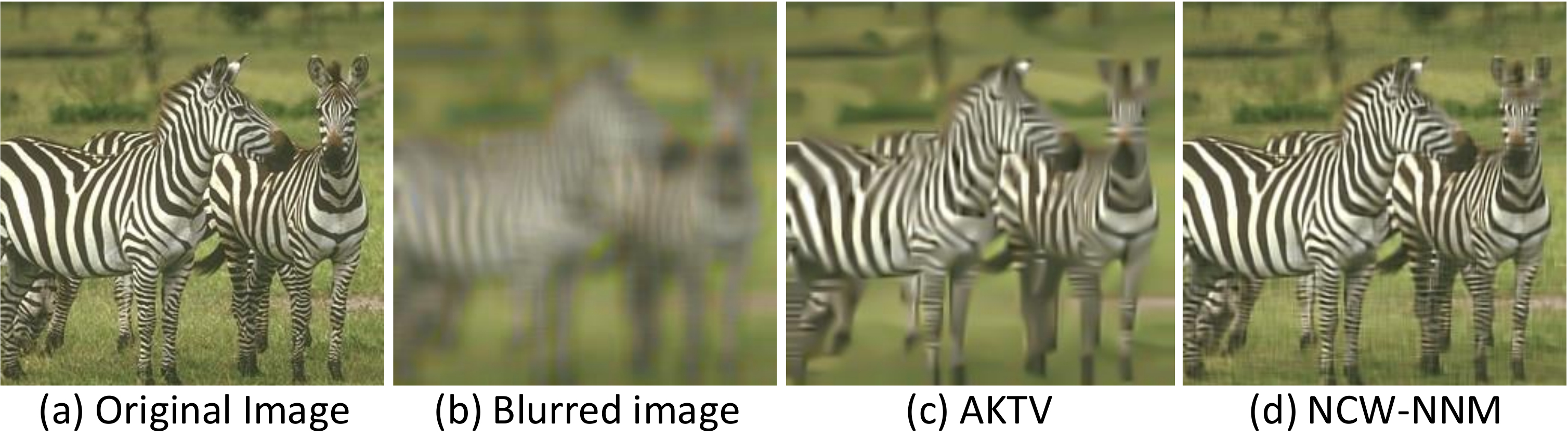}}
\end{minipage}
\caption{Deblurring performance comparison on the \emph{Zebra} image. (a) Original image; (b) Noisy and blurred image ($19\times 19$ uniform blur, BSNR=40); deblurred image by (c) AKTV \cite{62} (PSNR = 21.18dB, FSIM = 0.7394); (d) NCW-NNM (PSNR = \textbf{26.15dB}, FSIM = \textbf{0.9028}).}
\label{fig:4}
\end{figure*}

In addition, the proposed NCW-NNM is compared with AKTV method \cite{62}, where AKTV method is well-known for working quite well in the case of large blur.  Here, $19 \times 19$ uniform kernel with the corresponding BSNR=40 for image $\emph{Fence}$ is tested, where BSNR means Blurred Signal to Noise Ratio, and it is equivalent to 10*log (Blurred Signal variance/Noise variance). Smaller BSNR means larger noise variance. All the parameters are the same as specified in the case of $9\times 9$ uniform kernel except for $\rho=0.00002$. The visual quality comparison is shown in Fig.~\ref{fig:4}. It can be seen that the proposed NCW-NNM produces much clear and preserves much more details than AKTV method.

 \begin{table*}[!htbp]
\caption{PSNR/FSIM Comparisons for Image Deblurring.}
\centering  
\footnotesize
\begin{tabular}{|c|c|c|c|c|c|c|}
\hline
  \multicolumn{7}{|c|}{{{Gaussian Kernel: fspecial('gaussian', 25, 1.6), $\delta=\sqrt{2}$}}} \\
\hline
\multirow{1}{*}{\textbf{{Method}}}&{\emph{{Light}}}&{\emph{{Barbara}}}&{\emph{{Fence}}}
&{\emph{{Flower}}}&{\emph{{Zebra}}}&{\textbf{{Average}}}\\
 \hline
 \multirow{1}{*}{BM3D \cite{59}}& 21.75/0.8430 &  25.77/0.8802  & 27.31/0.8936 & 29.84/0.9100 & 24.64/0.8666 & 25.86/0.8787\\
\hline
 \multirow{1}{*}{NCSR \cite{26}}& \textbf{22.28}/0.8667 &  25.95/0.8853  & 27.40/0.9087 & 30.21/0.9187 & \textbf{25.05/0.8868} & 26.18/0.8932\\
\hline
 \multirow{1}{*}{JSM \cite{7}}& 22.24/0.8612 &  25.89/0.8752  & 27.08/0.9026 & 29.52/0.8878 & 24.66/0.8722 & 25.88/0.8798\\
\hline
 \multirow{1}{*}{MSEPLL \cite{60}}& 21.93/0.8634 &  24.82/0.8579  & 27.12/0.8933 & 30.35/0.9234 & 24.75/0.8864 & 25.79/0.8849\\
\hline
 \multirow{1}{*}{FPD \cite{61}}& 21.12/0.8319 &  24.54/0.8622  & 26.54/0.8886 & \textbf{30.46/0.9354} & 24.01/0.8716 & 25.33/0.8780\\
\hline
 \multirow{1}{*}{NNM }& 21.88/0.8496 &  25.90/0.8795  & 27.35/0.9074 & 29.94/0.9040 & 24.60/0.8721 & 25.93/0.8825\\
\hline
 \multirow{1}{*}{NCW-NNM }& 22.26/\textbf{0.8683} &  \textbf{26.72/0.9009}  & \textbf{27.57/0.9136} & {30.27/0.9230} & 24.88/0.8805 & \textbf{26.34/0.8973}\\
\hline
  \multicolumn{7}{|c|}{{{9$\times 9$ Uniform Kernel, $\delta=\sqrt{2}$}}} \\
\hline
\multirow{1}{*}{\textbf{{Method}}}&{\emph{{Light}}}&{\emph{{Barbara}}}&{\emph{{Fence}}}
&{\emph{{Flower}}}&{\emph{{Zebra}}}&{\textbf{{Average}}}\\
 \hline
 \multirow{1}{*}{BM3D \cite{59}}& 21.65/0.8423 &  26.89/0.8807  & 28.94/0.9045 & 28.54/0.8732 & 23.69/0.8330 & 25.94/0.8667\\
\hline
 \multirow{1}{*}{NCSR \cite{26}}& 22.10/0.8733 &  27.12/0.9006  & 29.83/\textbf{0.9302} & \textbf{29.28}/0.9001 & \textbf{24.63/0.8743} & 26.59/0.8957\\
\hline
 \multirow{1}{*}{JSM \cite{7}}& 21.84/0.8439  &  25.72/0.8600  & 27.26/0.9036 & 27.15/0.8257 & 23.32/0.8274 & 25.06/0.8521\\
\hline
 \multirow{1}{*}{MSEPLL \cite{60}}& 21.27/0.8417 &  26.24/0.8756  & 28.02/0.8935 & 29.07/0.8881 & 23.98/0.8566 & 25.72/0.8711\\
\hline
 \multirow{1}{*}{FPD \cite{61}}& 19.70/0.7725 &  25.22/0.8519  & 25.25/0.8390 & 28.49/0.8799 & 21.62/0.7882 & 24.06/0.8263\\
\hline
 \multirow{1}{*}{NNM }& 21.82/0.8579 &  27.02/0.8945  & 29.76/0.9284 & 28.91/0.8875 & 23.84/0.8541 & 26.27/0.8845\\
\hline
 \multirow{1}{*}{NCW-NNM}& \textbf{22.58/0.8790} &  \textbf{27.48/0.9092}  & \textbf{29.98}/0.9300 & 29.27/\textbf{0.9043} & 24.43/0.8682 & \textbf{26.75/0.8981}\\
\hline
\end{tabular}
\label{lab:2}
\end{table*}

\subsection{Image Inpainting}

In this subsection, two interesting examples of image inpainting with different masks are conducted, i.e., image restoration from partial random samples and text inlayed sample. The parameters are set as follows. The size of each patch $\sqrt{d}\times \sqrt{d}$ is set to be $8\times 8$ and $10\times 10$ for partial random samples and text inlayed, respectively. Similar patch numbers $m =60$, $\emph{L}=25$, $\epsilon=0.1$, $\varsigma=0.3$, $J=2$. ($\rho$, $p$) are set to (0.0003, 0.45), (0.0003, 0.45), (0.03, 0.95), (0.04, 0.95) and (0.06, 0.95) when 80\%, 70\%, 60\%, 50\% pixels missing and text inlayed, respectively.

\begin{figure*}[!htbp]
\begin{minipage}[b]{1\linewidth}
  \centerline{\includegraphics[width=12cm]{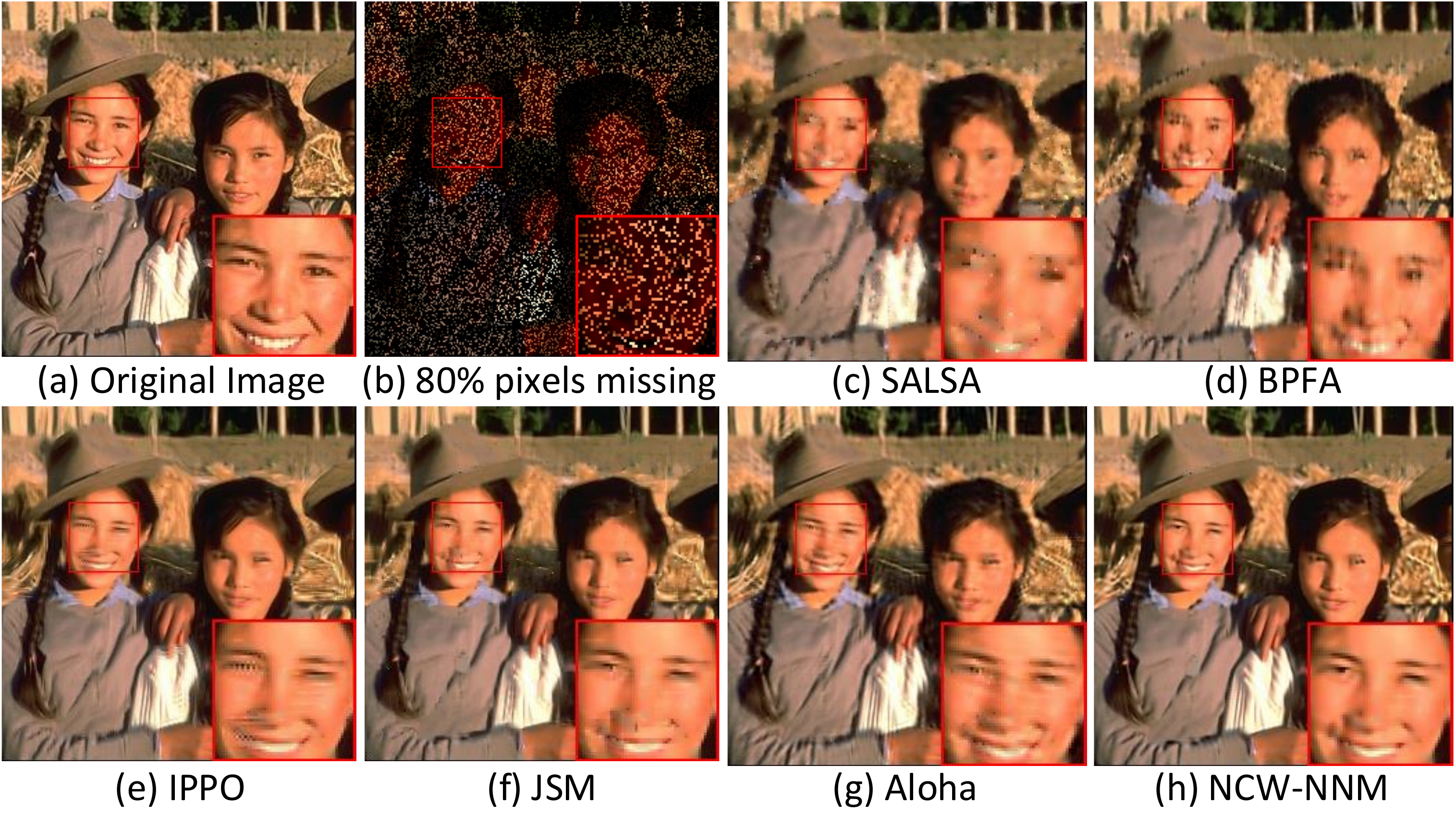}}
\end{minipage}
\caption{Visual comparison of \emph{Girl} by image inpainting with 80\% pixels missing. (a) Original image; (b) Degraded image with 80\% missing sample; Restoration by (c) SALSA \cite{69} (PSNR = 23.80dB, FSIM = 0.8517); (d) BPFA \cite{10} (PSNR = 24.80dB, FSIM = 0.8765); (e) IPPO \cite{63} (PSNR = 25.31dB, FSIM = 0.8914); (f) JSM \cite{7} (PSNR = 25.18dB, FSIM = 0.8871); (g) Aloha \cite{9} (PSNR = 25.17dB, FSIM = 0.8832); (h) NCW-NNM (PSNR = \textbf{25.78dB}, FSIM = \textbf{0.9065}).}
\label{fig:5}
\end{figure*}

\begin{figure*}[!htbp]
\begin{minipage}[b]{1\linewidth}
  \centerline{\includegraphics[width=12cm]{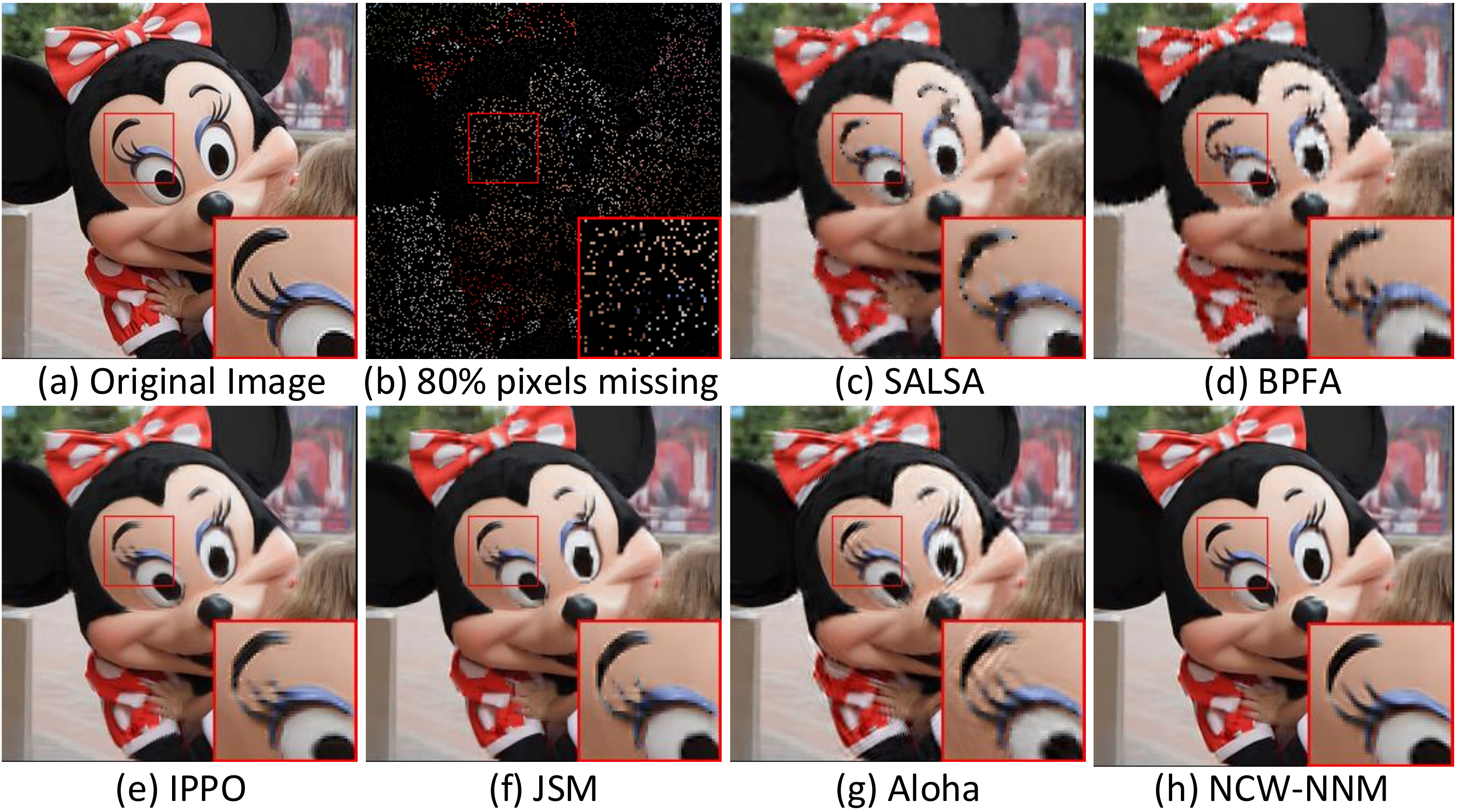}}
\end{minipage}
\caption{Visual comparison of \emph{Mickey} by image inpainting with 80\% pixels missing. (a) Original image; (b) Degraded image with 80\% missing sample; Restoration by (c) SALSA \cite{69} (PSNR = 24.46dB, FSIM = 0.8693); (d) BPFA \cite{10} (PSNR = 24.53dB, FSIM = 0.8696); (e) IPPO \cite{63} (PSNR = 26.33dB, FSIM = 0.9099); (f) JSM \cite{7} (PSNR = 26.09dB, FSIM = 0.9060); (g) Aloha \cite{9} (PSNR = 25.33dB, FSIM = 0.8770); (h) NCW-NNM (PSNR = \textbf{26.92dB}, FSIM = \textbf{0.9214}).}
\label{fig:6}
\end{figure*}

We compare the proposed NCW-NNM  with five other competing methods: SALSA \cite{69}, BPFA \cite{10}, IPPO \cite{63}, JSM \cite{7} and  Aloha \cite{9}. Table~\ref{lab:3} lists the PSNR and FSIM comparison results. In terms of PSNR, the proposed NCW-NNM achieves 3.52dB, 2.26dB, 1.10dB, 1.27dB and 1.98dB improvement on average over the SALSA, BPFA, IPPO, JSM and Aloha, respectively. The visual comparisons of the image inpainting methods are shown in Fig.~\ref{fig:5} and Fig.~\ref{fig:6}. It can be seen that SALSA, BPFA, IPPO, JSM and Aloha still generate some undesirable ringing effects and some details are lost. In contract, the proposed NCW-NNM  not only preserves sharper edges and finer details, but also significantly eliminates the ringing effects.
 \begin{table*}[!htbp]
\caption{PSNR/FSIM Comparisons for Image Inpainting.}
\centering  
\scriptsize
\begin{tabular}{|c|c|c|c|c|c|c|}
\hline
  \multicolumn{7}{|c|}{\textbf{{80\% pixels missing}}} \\
\hline
\multirow{1}{*}{\textbf{{Method}}}&{\emph{{Mickey}}}&{\emph{{Butterfly}}}&{\emph{{Castle}}}
&{\emph{{Corn}}}&{\emph{{Girl}}}&{\textbf{{Average}}}\\
 \hline
 \multirow{1}{*}{SALSA \cite{69} } & 24.46/0.8693 &  22.85/0.8451  & 22.85/0.8319 & 24.28/0.8803 & 23.80/0.8517 & 23.65/0.8557\\
\hline
 \multirow{1}{*}{BPFA \cite{10}}   & 24.53/0.8696 &  24.04/0.8532  & 23.94/0.8639 & 25.54/0.9010 & 24.80/0.8765 & 24.57/0.8729\\
\hline
 \multirow{1}{*}{IPPO \cite{63} }  & 26.33/0.9099 &  25.13/0.9078  & 24.50/0.8818 & 25.14/0.9020 & 25.31/0.8914 & 25.28/0.8986\\
\hline
 \multirow{1}{*}{JSM  \cite{7}}   & 26.09/0.9060 &  25.57/0.9125  & 24.59/0.8830 & 25.58/0.9089 & 25.18/0.8871 & 25.40/0.8995\\
\hline
 \multirow{1}{*}{Aloha \cite{9}}  & 25.33/0.8770 &  24.88/0.8586  & 23.88/0.8728 & 25.60/0.8963 & 25.16/0.8832 & 24.97/0.8775\\
\hline
 \multirow{1}{*}{NNM }   & 25.97/0.9011 &  25.61/0.9125  & 24.35/0.8689 & 25.68/0.9076 & 25.18/0.8824 & 25.36/0.8945\\
\hline
 \multirow{1}{*}{NCW-NNM } & \textbf{26.92/0.9214} &  \textbf{26.52/0.9271}  & \textbf{24.84/0.8953} & \textbf{26.87/0.9287} & \textbf{25.78/0.9065} & \textbf{26.19/0.9158}\\
\hline
  \multicolumn{7}{|c|}{\textbf{{70\% pixels missing}}} \\
\hline
\multirow{1}{*}{\textbf{{Method}}}&{\emph{{Mickey}}}&{\emph{{Butterfly}}}&{\emph{{Castle}}}
&{\emph{{Corn}}}&{\emph{{Girl}}}&{\textbf{{Average}}}\\
 \hline
 \multirow{1}{*}{SALSA \cite{69}}  & 25.98/0.9033 &  25.06/0.8909  & 24.22/0.8761 & 26.11/0.9193 & 25.48/0.8932 & 25.37/0.8965\\
\hline
 \multirow{1}{*}{BPFA \cite{10} }  & 26.16/0.9058 &  26.68/0.9077  & 25.66/0.9048 & 27.82/0.9366 & 26.86/0.9157 & 26.64/0.9141\\
\hline
 \multirow{1}{*}{IPPO \cite{63} }  & 28.59/0.9406 &  27.69/0.9401  & 26.11/0.9162 & 27.77/0.9409 & 27.43/0.9316 & 27.52/0.9339\\
\hline
 \multirow{1}{*}{JSM \cite{7}}    & 28.25/0.9356 &  27.97/0.9430  & 26.64/0.9217 & 27.67/0.9412 & 27.20/0.9275 & 27.55/0.9338\\
\hline
 \multirow{1}{*}{Aloha \cite{9}}  & 27.11/0.9107 &  27.29/0.8996  & 25.77/0.9101 & 27.95/0.9314 & 27.08/0.9211 & 27.04/0.9146\\
\hline
 \multirow{1}{*}{NNM}       & 27.86/0.9311 &  27.86/0.9414  & 26.15/0.9108 & 27.63/0.9390 & 26.92/0.9209 & 27.28/0.9286\\
\hline
 \multirow{1}{*}{NCW-NNM }  & \textbf{29.29/0.9474} &  \textbf{29.28/0.9532}  & \textbf{26.87/0.9295} & \textbf{29.29/0.9562} & \textbf{28.09/0.9419} & \textbf{28.56/0.9456}\\
\hline
  \multicolumn{7}{|c|}{\textbf{{60\% pixels missing}}} \\
\hline
\multirow{1}{*}{\textbf{{Method}}}&{\emph{{Mickey}}}&{\emph{{Butterfly}}}&{\emph{{Castle}}}
&{\emph{{Corn}}}&{\emph{{Girl}}}&{\textbf{{Average}}}\\
 \hline
 \multirow{1}{*}{SALSA \cite{69} } & 27.41/0.9262 &  26.79/0.9199  & 25.73/0.9093 & 27.75/0.9438 & 27.02/0.9244 & 26.94/0.9247\\
\hline
 \multirow{1}{*}{BPFA \cite{10}}   & 23.65/0.9186 &  28.88/0.9394  & 27.28/0.9313 & 30.07/0.9590 & 28.75/0.9417 & 27.73/0.9380\\
\hline
 \multirow{1}{*}{IPPO \cite{63}}   & 30.77/0.9592 &  29.85/0.9600  & 27.81/0.9396 & 29.75/0.9610 & 29.32/0.9528 & 29.50/0.9545\\
\hline
 \multirow{1}{*}{JSM \cite{7}}    & 29.85/0.9536 &  29.83/0.9600  & 28.09/0.9426 & 29.45/0.9610 & 29.01/0.9501 & 29.25/0.9535\\
\hline
 \multirow{1}{*}{Aloha \cite{9}}  & 28.59/0.9342 &  29.16/0.9242  & 27.16/0.9298 & 29.83/0.9526 & 28.91/0.9434 & 28.73/0.9368\\
\hline
 \multirow{1}{*}{NNM}       & 29.61/0.9508 &  29.82/0.9595  & 27.61/0.9359 & 29.33/0.9587 & 28.71/0.9462 & 29.02/0.9500\\
\hline
 \multirow{1}{*}{NCW-NNM }  & \textbf{31.46/0.9643} &  \textbf{31.54/0.9690}  & \textbf{28.44/0.9484} & \textbf{31.52/0.9724} & \textbf{30.23/0.9624} & \textbf{30.64/0.9633}\\
\hline
  \multicolumn{7}{|c|}{\textbf{{50\% pixels missing}}}\\
\hline
\multirow{1}{*}{\textbf{{Method}}}&{\emph{{Mickey}}}&{\emph{{Butterfly}}}&{\emph{{Castle}}}
&{\emph{{Corn}}}&{\emph{{Girl}}}&{\textbf{{Average}}}\\
 \hline
 \multirow{1}{*}{SALSA \cite{69}} & 28.98/0.9458 &  28.52/0.9432  & 27.01/0.9314 & 29.39/0.9600 & 28.60/0.9455 & 28.50/0.9452\\
\hline
 \multirow{1}{*}{BPFA \cite{10}}  & 29.43/0.9501 &  30.98/0.9595  & 28.83/0.9488 & 32.10/0.9725 & 30.58/0.9597 & 30.38/0.9581\\
\hline
 \multirow{1}{*}{IPPO \cite{63}}  & 32.74/0.9719 &  31.69/0.9724  & 29.57/0.9576 & 31.76/0.9745 & 31.05/0.9672 & 31.36/0.9687\\
\hline
 \multirow{1}{*}{JSM \cite{7}}   & 31.96/0.9685 &  31.47/0.9720  & 29.48/0.9583 & 31.33/0.9743 & 30.68/0.9665 & 30.98/0.9679\\
\hline
 \multirow{1}{*}{Aloha \cite{9}} & 30.33/0.9515 &  30.78/0.9414  & 28.71/0.9485 & 31.89/0.9679 & 30.60/0.9608 & 30.46/0.9540\\
\hline
 \multirow{1}{*}{NNM}      & 31.62/0.9666 &  31.46/0.9715  & 29.17/0.9540 & 31.23/0.9731 & 30.40/0.9641 & 30.77/0.9659\\
\hline
 \multirow{1}{*}{NCW-NNM } & \textbf{34.01/0.9769} &  \textbf{33.26/0.9785}  & \textbf{30.15/0.9634} & \textbf{33.80/0.9828} & \textbf{32.13/0.9749} & \textbf{32.67/0.9753}\\
\hline
  \multicolumn{7}{|c|}{\textbf{{Inlay text}}} \\
\hline
\multirow{1}{*}{\textbf{{Method}}}&{\emph{{Mickey}}}&{\emph{{Butterfly}}}&{\emph{{Castle}}}
&{\emph{{Corn}}}&{\emph{{Girl}}}&{\textbf{{Average}}}\\
 \hline
 \multirow{1}{*}{SALSA \cite{69}} & 30.67/0.9670 &  29.81/0.9617  & 28.75/0.9556 & 30.96/0.9736 & 30.27/0.9630 & 30.09/0.9642\\
\hline
 \multirow{1}{*}{BPFA \cite{10}}  & 31.71/0.9719 &  31.71/0.9683  & 30.94/0.9671 & 32.16/0.9782 & 31.28/0.9691 & 31.56/0.9710\\
\hline
 \multirow{1}{*}{IPPO \cite{63}}  & 34.04/0.9838 &  33.98/0.9840  & 31.91/0.9762 & 32.48/0.9815 & 32.65/0.9795 & 33.01/0.9810\\
\hline
 \multirow{1}{*}{JSM \cite{7}}   & 32.99/0.9811 &  33.19/0.9831  & 32.48/0.9769 & 32.26/0.9815 & 32.21/0.9772 & 32.62/0.9799\\
\hline
 \multirow{1}{*}{Aloha \cite{9}} & 30.49/0.9641 &  31.58/0.9569  & 30.34/0.9672 & 32.04/0.9752 & 30.84/0.9693 & 31.06/0.9665\\
\hline
 \multirow{1}{*}{ NNM}   & 32.65/0.9794 &  33.00/0.9805  & 31.49/0.9727 & 32.15/0.9802 & 32.07/0.9753 & 32.27/0.9776\\
\hline
 \multirow{1}{*}{NCW-NNM } & \textbf{35.31/0.9869} &  \textbf{34.89/0.9861}  & \textbf{33.17/0.9809} & \textbf{33.98/0.9870} & \textbf{33.20/0.9818} & \textbf{34.11/0.9845}\\
\hline
\end{tabular}
\label{lab:3}
\end{table*}

\subsection{Image Compressive Sensing Recovery}

In this subsection, we show the experimental results of the proposed NCW-NNM based image CS recovery. We generate the CS measurements at the block level by exploiting Gaussian random projection matrix to test images, i.e., the block-based CS recovery with block size of $32 \times 32$. The parameters are set as follows. The size of each patch $\sqrt{d} \times \sqrt{d}$ is set to be $7 \times 7$.  The searching matched patches $m$ = 60 and $L=20$. We assign $\epsilon=0.1$ and $\varsigma=0.4$.  (p, $\rho$) are set to (0.65, 0.0001), (0.5, 0.0005), (0.95, 0.005) and (0.95, 0.005) when 0.1$N$, 0.2$N$, 0.3$N$ and 0.4$N$ measurements, respectively.
\begin{figure*}[!htbp]
\begin{minipage}[b]{1\linewidth}
  \centerline{\includegraphics[width=12cm]{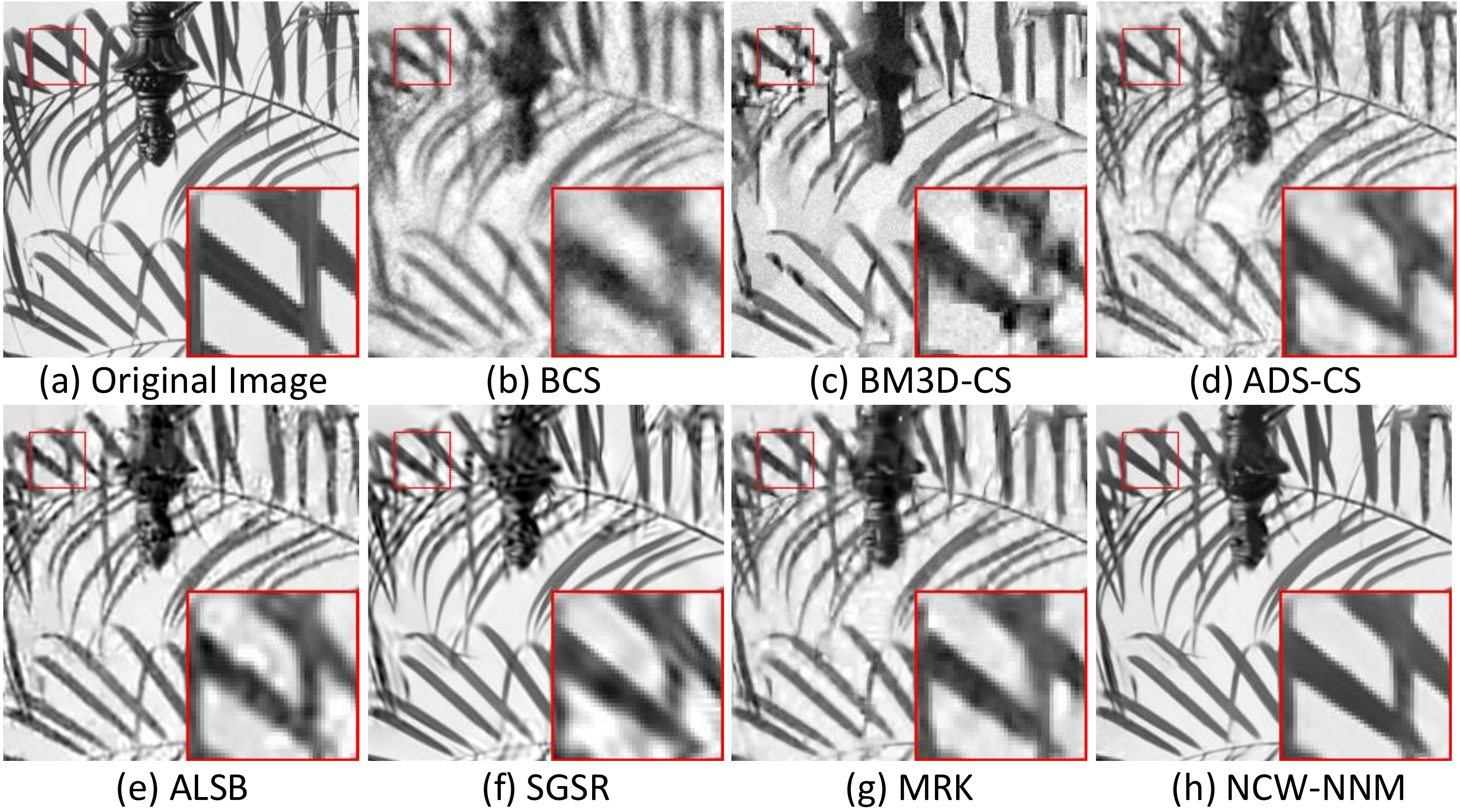}}
\end{minipage}
\caption{Visual comparison of \emph{Leaves} by image CS recovery with 0.1$N$ measurements. (a) Original image; (b) BCS \cite{64} (PSNR =18.55dB, FSIM = 0.5797); (c) BM3D-CS \cite{65} (PSNR = 18.93dB, FSIM = 0.7259); (d) ADS-CS \cite{15} (PSNR = 21.24dB, FSIM = 0.7671); (e) ALSB \cite{23} (PSNR = 21.61dB, FSIM = 0.8010); (f) SGSR \cite{66} (PSNR = 22.22dB, FSIM = 0.8356); (g) MRK \cite{67} (PSNR = 22.05dB, FSIM = 0.8118); (h) NCW-NNM (PSNR = \textbf{22.05dB}, FSIM = \textbf{0.8118}).}
\label{fig:7}
\end{figure*}

\begin{figure*}[!htbp]
\begin{minipage}[b]{1\linewidth}
  \centerline{\includegraphics[width=12cm]{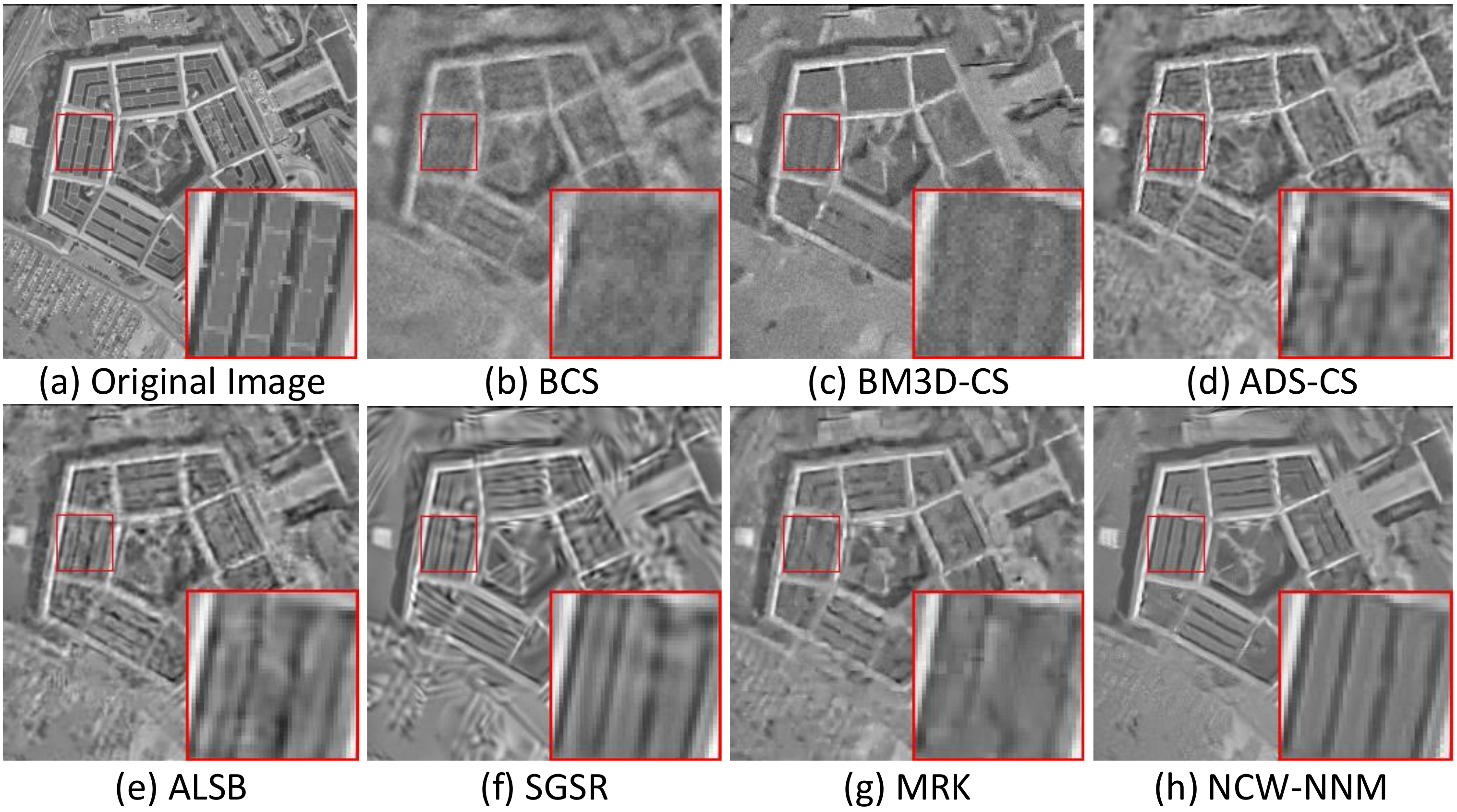}}
\end{minipage}
\caption{Visual comparison of \emph{pentagon} by image CS recovery with 0.1$N$ measurements. (a) Original image; (b) BCS \cite{64} (PSNR =22.14dB, FSIM = 0.5068); (c) BM3D-CS \cite{65} (PSNR = 21.87dB, FSIM = 0.7443); (d) ADS-CS \cite{15} (PSNR = 23.37dB, FSIM = 0.7999); (e) ALSB \cite{23} (PSNR = 23.16dB, FSIM =0.8059); (f) SGSR \cite{66} (PSNR = 23.05dB, FSIM = 0.8060); (g) MRK \cite{67} (PSNR = 23.70dB, FSIM = 0.7890); (h) NCW-NNM (PSNR = \textbf{24.26dB}, FSIM = \textbf{0.8064}).}
\label{fig:8}
\end{figure*}

To verify the performance of the proposed NCW-NNM, we have compared it with some competitive CS recovery methods including BCS \cite{64}, BM3D-CS \cite{65}, ADS-CS \cite{15}, ALSB \cite{23}, SGSR \cite{66} and MRK \cite{67} methods. The PSNR and FSIM results of 6 gray images are shown in Table~\ref{lab:4}. One can observe that the proposed NCW-NNM almost outperforms other competing methods on all test images over different numbers of CS measurements. The average gain of NCW-NNM over BCS, BM3D-CS, ADS-CS, ALSB, SGSR and MRK can be as much as 6.48dB, 2.94dB, 1.28dB, 1.77dB, 1.76dB and 1.62dB, respectively. The visual comparison results of the recovered images are presented in Fig.~\ref{fig:7} and Fig.~\ref{fig:8}. It can be seen that BCS method generates the worst perceptual result. The BM3D-CS, ADS-CS, ALSB, SGSR and MRK methods can obtain much better visual quality than BCS method, but still suffer from some undesirable artifacts or over-smooth phenomena, such as ring effects and some fine image details are lost. By contrast, the proposed NCW-NNM not only removes most of the visual artifacts, but also preserves large-scale sharp edges and small-scale fine image details more effective than other competing methods.
 \begin{table*}[!htbp]
\caption{PSNR/FSIM Comparisons for Image CS Recovery.}
\centering  
\tiny
\begin{tabular}{|c|c|c|c|c|c|c|c|}
\hline
  \multicolumn{8}{|c|}{Subrate =0.1$N$} \\
\hline
\multirow{1}{*}{\textbf{{Method}}}&{\emph{{boats}}}&{\emph{{House}}}&{\emph{{fingerprint}}}
&{\emph{{Leaves}}}&{\emph{{pentagon}}}&{\emph{{starfish}}} &{\textbf{{Average}}}\\
 \hline
 \multirow{1}{*}{BCS \cite{64} }        & 24.54/0.6715 &  26.99/0.7681  & 17.10/0.3434 & 18.55/0.5797 & 22.14/0.5068 & 22.97/0.6687 & 22.05/0.5897\\
\hline
 \multirow{1}{*}{BM3D-CS \cite{65} }    & 25.40/0.8308 &  32.50/0.9146  & 16.02/0.6684 & 18.93/0.7259 & 21.87/0.7443 & 20.72/0.7591 & 22.57/0.7739\\
\hline
 \multirow{1}{*}{ADS-CS \cite{15} }     & 28.30/0.8886 &  33.39/0.9057  & 18.91/0.7714 & 21.24/0.7671 & 23.37/0.7999 & \textbf{25.50}/0.8579 & 25.12/0.8318\\
\hline
 \multirow{1}{*}{ALSB \cite{23} }       & 28.09/0.8894 &  32.17/0.9107  & 20.68/0.8665 & 21.61/0.8010 & 23.16/0.8059 & 23.61/0.8305 & 24.89/0.8507\\
\hline
 \multirow{1}{*}{SGSR \cite{66} }       & 27.71/0.8915 &  32.77/0.9187  & 20.37/0.8671 & 22.22/0.8356 & 23.05/0.8060 & 22.91/0.8175 & 24.84/0.8560\\
\hline
 \multirow{1}{*}{MRK \cite{67} }        & 28.32/0.8906 &  32.43/0.9132  & 17.66/0.6774 & 22.05/0.8118 & 23.70/0.7890 & 25.27/0.8580 & 24.90/0.8233\\
\hline
 \multirow{1}{*}{NNM  }        & 27.91/0.8925 &  32.89/0.9260  & 21.33/0.8558 & 22.85/0.8880 & 24.02/0.7961 & 24.21/0.8484 & 25.70/0.8678\\
\hline
 \multirow{1}{*}{NCW-NNM  }    & \textbf{28.64/0.9019} &  \textbf{33.52/0.9279}  & \textbf{21.57/0.8684} &  \textbf{25.77/0.9124} & \textbf{24.26/0.8064} & 25.49/\textbf{0.8755} & \textbf{26.54/0.8821}\\
\hline
  \multicolumn{8}{|c|}{Subrate =0.2$N$} \\
\hline
\multirow{1}{*}{\textbf{{Method}}}&{\emph{{boats}}}&{\emph{{House}}}&{\emph{{fingerprint}}}
&{\emph{{Leaves}}}&{\emph{{pentagon}}}&{\emph{{starfish}}} &{\textbf{{Average}}}\\
 \hline
 \multirow{1}{*}{BCS \cite{64} }        & 27.05/0.8654 &  30.54/0.9011  & 18.50/0.7355 & 21.12/0.7531 & 23.97/0.8087 & 25.29/0.8624 & 24.41/0.8211\\
\hline
 \multirow{1}{*}{BM3D-CS \cite{65} }    & 31.01/0.9314 &  35.04/0.9498  & 19.38/0.8184 & 28.14/0.9232 & 25.49/0.8558 & 27.50/0.8964 & 27.76/0.8958\\
\hline
 \multirow{1}{*}{ADS-CS \cite{15} }     & 33.15/0.9508 &  35.76/0.9423  & 22.70/0.8976 & 27.88/0.9015 & 26.31/0.8801 & \textbf{30.22}/0.9262 & 29.34/0.9164\\
\hline
 \multirow{1}{*}{ALSB \cite{23} }       & 32.96/0.9514 &  36.07/0.9563  & 23.69/0.9226 & 27.15/0.9089 & 26.19/0.8817 & 27.30/0.8984 & 28.89/0.9199\\
\hline
 \multirow{1}{*}{SGSR \cite{66} }       & 32.41/0.9466 &  35.81/0.9503  & 23.20/0.9186 & 28.74/0.9373 & 26.55/0.8947 & 27.13/0.8986 & 28.97/0.9243\\
\hline
 \multirow{1}{*}{MRK \cite{67} }        & 32.38/0.9476 &  36.36/0.9586  & 20.54/0.8397 & 27.75/0.9169 & 27.11/0.8888 & 29.18/0.9239 & 28.89/0.9126\\
\hline
 \multirow{1}{*}{NNM  }        & 31.68/0.9422 &  36.01/0.9580  & 23.57/0.9110 & 27.72/0.9373 & 26.60/0.8806 & 27.37/0.9092 & 28.83/0.9230\\
\hline
 \multirow{1}{*}{NCW-NNM }    & \textbf{33.77/0.9576} &  \textbf{36.92/0.9616}  & \textbf{24.20/0.9287} &  \textbf{31.46/0.9615} & \textbf{27.59/0.9072} & 29.99/\textbf{0.9398} & \textbf{30.65/0.9427}\\
\hline
  \multicolumn{8}{|c|}{Subrate =0.3$N$} \\
\hline
\multirow{1}{*}{\textbf{{Method}}}&{\emph{{boats}}}&{\emph{{House}}}&{\emph{{fingerprint}}}
&{\emph{{Leaves}}}&{\emph{{pentagon}}}&{\emph{{starfish}}} &{\textbf{{Average}}}\\
 \hline
 \multirow{1}{*}{BCS \cite{64} }        & 28.91/0.8997 &  32.85/0.9299  & 19.96/0.8149 & 23.16/0.8018 & 25.54/0.8595 & 27.20/0.8968 & 26.27/0.8671\\
\hline
 \multirow{1}{*}{BM3D-CS \cite{65} }    & 34.04/0.9630 &  36.84/0.9689  & 23.02/0.9111 & 32.52/0.9602 & 28.20/0.9154 & 31.48/0.9444 & 31.02/0.9438\\
\hline
 \multirow{1}{*}{ADS-CS \cite{15} }     & 36.35/0.9728 &  38.21/0.9667  & 25.33/0.9408 & 32.55/0.9550 & 28.52/0.9217 & 32.90/0.9540 & 32.31/0.9518\\
\hline
 \multirow{1}{*}{ALSB \cite{23} }       & 36.42/0.9746 &  38.34/0.9732  & 25.84/0.9475 & 31.08/0.9511 & 28.22/0.9210 & 30.35/0.9408 & 31.71/0.9513\\
\hline
 \multirow{1}{*}{SGSR \cite{66} }       & 35.21/0.9683 &  37.37/0.9648  & 25.49/0.9455 & 32.98/0.9676 & 28.66/0.9318 & 30.78/0.9446 & 31.75/0.9538\\
\hline
 \multirow{1}{*}{MRK \cite{67} }        & 34.97/0.9687 &  38.35/0.9727  & 24.21/0.9225 & 32.37/0.9598 & 29.50/0.9337 & 32.53/0.9589 & 31.99/0.9527\\
\hline
 \multirow{1}{*}{NNM  }        & 34.18/0.9641 &  37.94/0.9721  & 25.27/0.9379 & 30.99/0.9619 & 28.41/0.9194 & 30.07/0.9422 & 31.14/0.9496\\
\hline
 \multirow{1}{*}{NCW-NNM  }    & \textbf{37.11/0.9772} &  \textbf{39.23/0.9766}  & \textbf{26.48/0.9544} &  \textbf{35.17/0.9798} & \textbf{29.74/0.9429} & \textbf{33.38/0.9654} & \textbf{33.52/0.9661}\\
\hline
  \multicolumn{8}{|c|}{Subrate =0.4$N$} \\
\hline
\multirow{1}{*}{\textbf{{Method}}}&{\emph{{boats}}}&{\emph{{House}}}&{\emph{{fingerprint}}}
&{\emph{{Leaves}}}&{\emph{{pentagon}}}&{\emph{{starfish}}} &{\textbf{{Average}}}\\
 \hline
 \multirow{1}{*}{BCS \cite{64} }        & 30.56/0.9248 &  34.65/0.9490  & 21.67/0.8747 & 25.07/0.8422 & 27.00/0.8950 & 28.94/0.9217 & 27.98/0.9012\\
\hline
 \multirow{1}{*}{BM3D-CS \cite{65} }    & 36.71/0.9805 &  38.08/0.9781  & 25.47/0.9451 & 36.01/0.9816 & 30.53/0.9483 & 34.27/0.9675 & 33.51/0.9669\\
\hline
 \multirow{1}{*}{ADS-CS \cite{15} }     & 38.79/0.9835 &  40.30/0.9803  & 27.32/0.9608 & 35.94/0.9763 & 30.64/0.9488 & 35.36/0.9721 & 34.73/0.9703\\
\hline
 \multirow{1}{*}{ALSB \cite{23} }       & 38.92/0.9840 &  40.25/0.9824  & 27.70/0.9638 & 34.57/0.9738 & 29.84/0.9422 & 32.98/0.9619 & 34.04/0.9680\\
\hline
 \multirow{1}{*}{SGSR \cite{66} }       & 37.41/0.9794 &  38.99/0.9759  & 27.64/0.9645 & 35.83/0.9799 & 30.66/0.9548 & 33.66/0.9661 & 34.03/0.9701\\
\hline
 \multirow{1}{*}{MRK \cite{67} }        & 37.20/0.9802 &  40.04/0.9819  & 26.83/0.9539 & 35.53/0.9783 & 31.49/0.9580 & 35.01/0.9745 & 34.35/0.9711\\
\hline
 \multirow{1}{*}{NNM  }        & 36.49/0.9768 &  39.70/0.9807  & 26.99/0.9568 & 33.72/0.9764 & 30.08/0.9441 & 32.51/0.9620 & 33.25/0.9662\\
\hline
 \multirow{1}{*}{NCW-NNM }    & \textbf{39.39/0.9855} &  \textbf{40.93/0.984}  & \textbf{28.67/0.971} &  \textbf{38.39/0.9887} & \textbf{31.94/0.9637} & \textbf{36.09/0.9785} & \textbf{35.90/0.9786}\\
\hline
\end{tabular}
\label{lab:4}
\end{table*}
\begin{figure*}[!htbp]
\begin{minipage}[b]{1\linewidth}
  \centerline{\includegraphics[width=12cm]{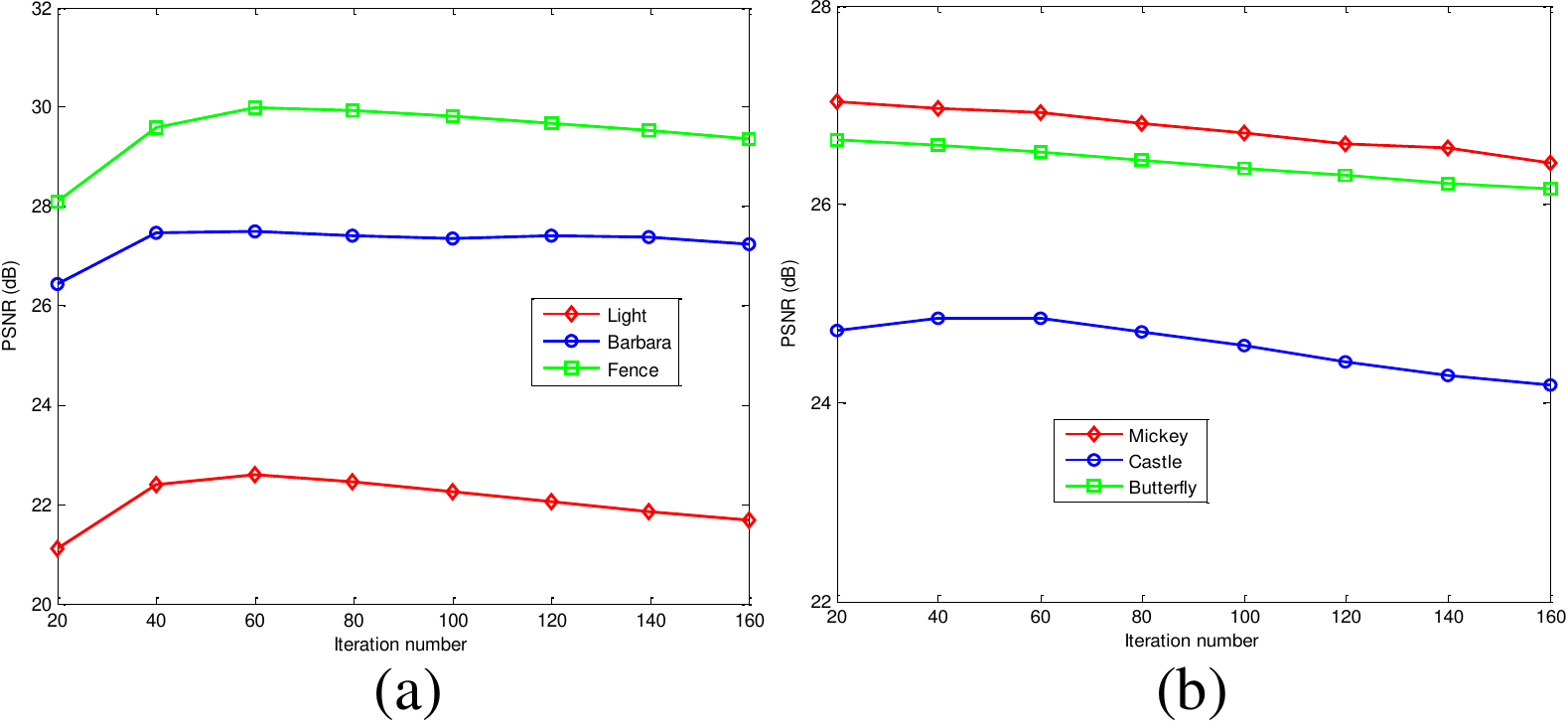}}
\end{minipage}
\caption{Performance comparison with different matched patch numbers $m$ for image deblurring and image inpainting. (a) PSNR results achieved by different $k$ in the case of $9\times 9$ Uniform Kernel, $\delta=\sqrt{2}$. (b)  PSNR results achieved by different $k$ in the case of the image inpainting with 80\% missing sample.}
\label{fig:9}
\end{figure*}
\subsection{Effect of the number of the best matched patches}
In this subsection, we have discussed how to select the best matching patch numbers $m$ for the performance of the proposed NCW-NNM method. Specifically, to investigate the sensitivity of our method against $m$, two experiments are conducted with respect to different $m$, ranging from 20 to 160, in the case of image deblurring and image inpainting, respectively. The results with different $m$ are shown in Fig.~\ref{fig:9}. It can be seen that all the curves are  almost flat, showing the performance of the proposed NCW-NNM is insensitive to $m$. The best performance of each case is usually achieved with $m$ in the range [40,80]. Therefore, in this work $m$ is empirically set to be 60.
\begin{figure*}[!htbp]
\begin{minipage}[b]{1\linewidth}
  \centerline{\includegraphics[width=12cm]{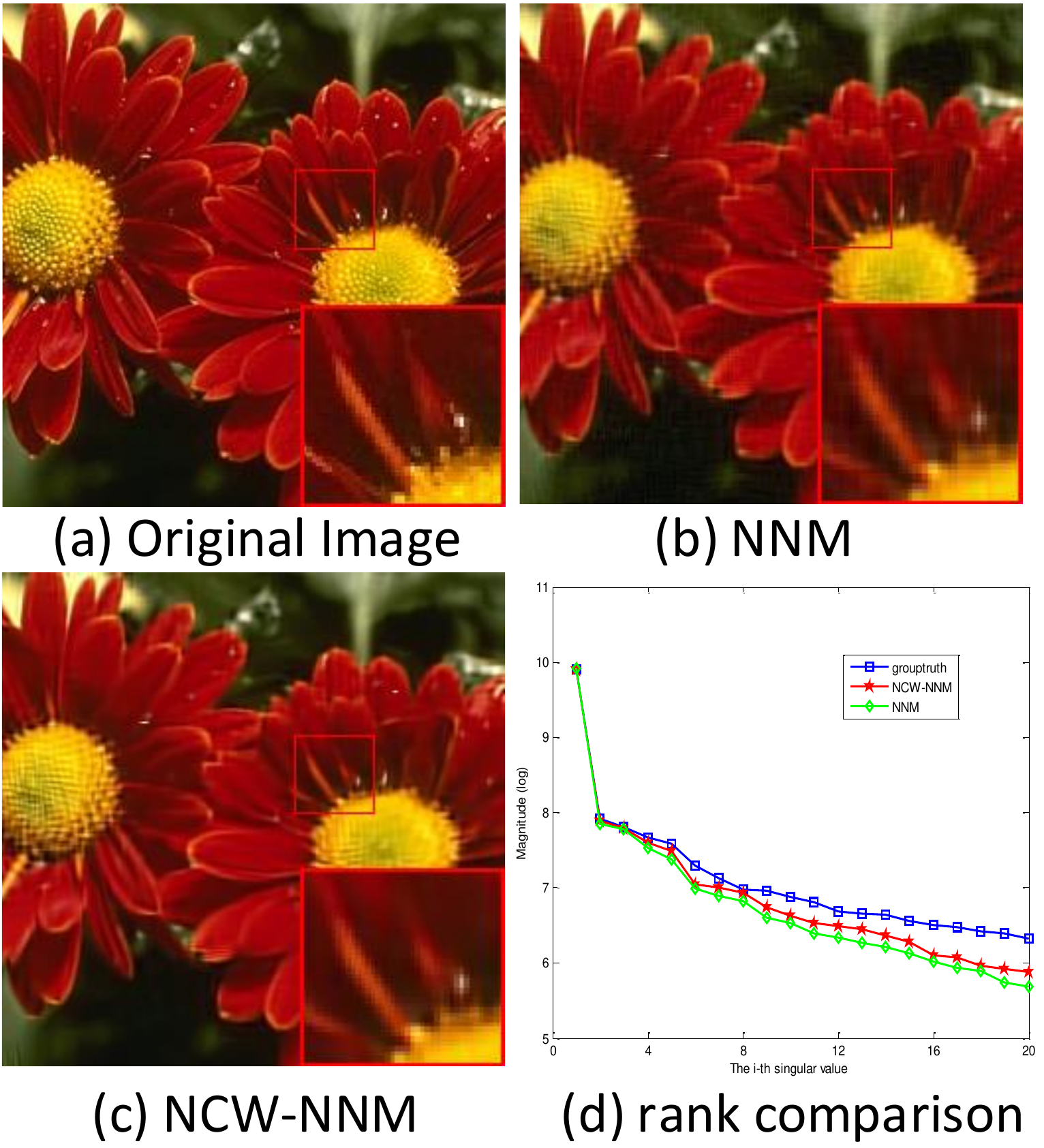}}
\end{minipage}
\caption{Deblurring performance comparison on the \emph{Flower} image with $19\times 19$ uniform blur, $\delta=\sqrt{2}$. (a) Original image; deblurred image by (b) NNM  (PSNR = 28.91dB, FSIM = 0.8875); (c) NCW-NNM (PSNR = \textbf{29.27dB}, FSIM = \textbf{0.9043}); (d) Comparison the rank of NNM and NCW-NNM methods.}
\label{fig:10}
\end{figure*}
\subsection{Comparison of NNM method}
\begin{figure*}[!htbp]
\begin{minipage}[b]{1\linewidth}
  \centerline{\includegraphics[width=12cm]{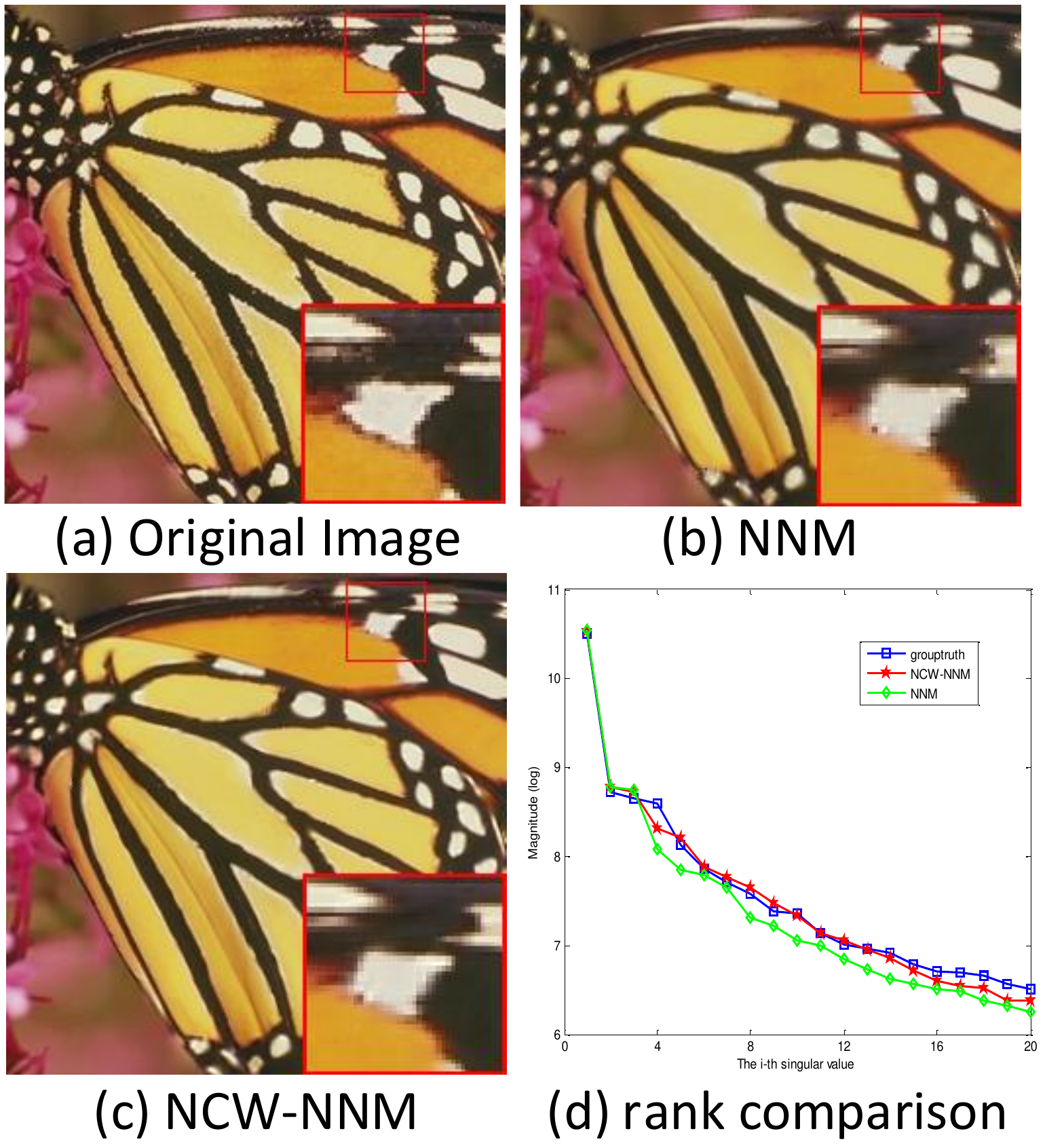}}
\end{minipage}
\caption{Visual comparison of \emph{Butterfly} by image inpainting with 80\% pixels missing. (a) Original image; (b) Degraded image with 80\% missing sample; deblurred image by (c) NNM  (PSNR = 25.61dB, FSIM = 0.9125); (d) NCW-NNM (PSNR = \textbf{26.52dB}, FSIM = \textbf{0.9271}); (e) Comparison the rank of NNM and NCW-NNM methods.}
\label{fig:11}
\end{figure*}
\begin{figure*}[!htbp]
\begin{minipage}[b]{1\linewidth}
  \centerline{\includegraphics[width=12cm]{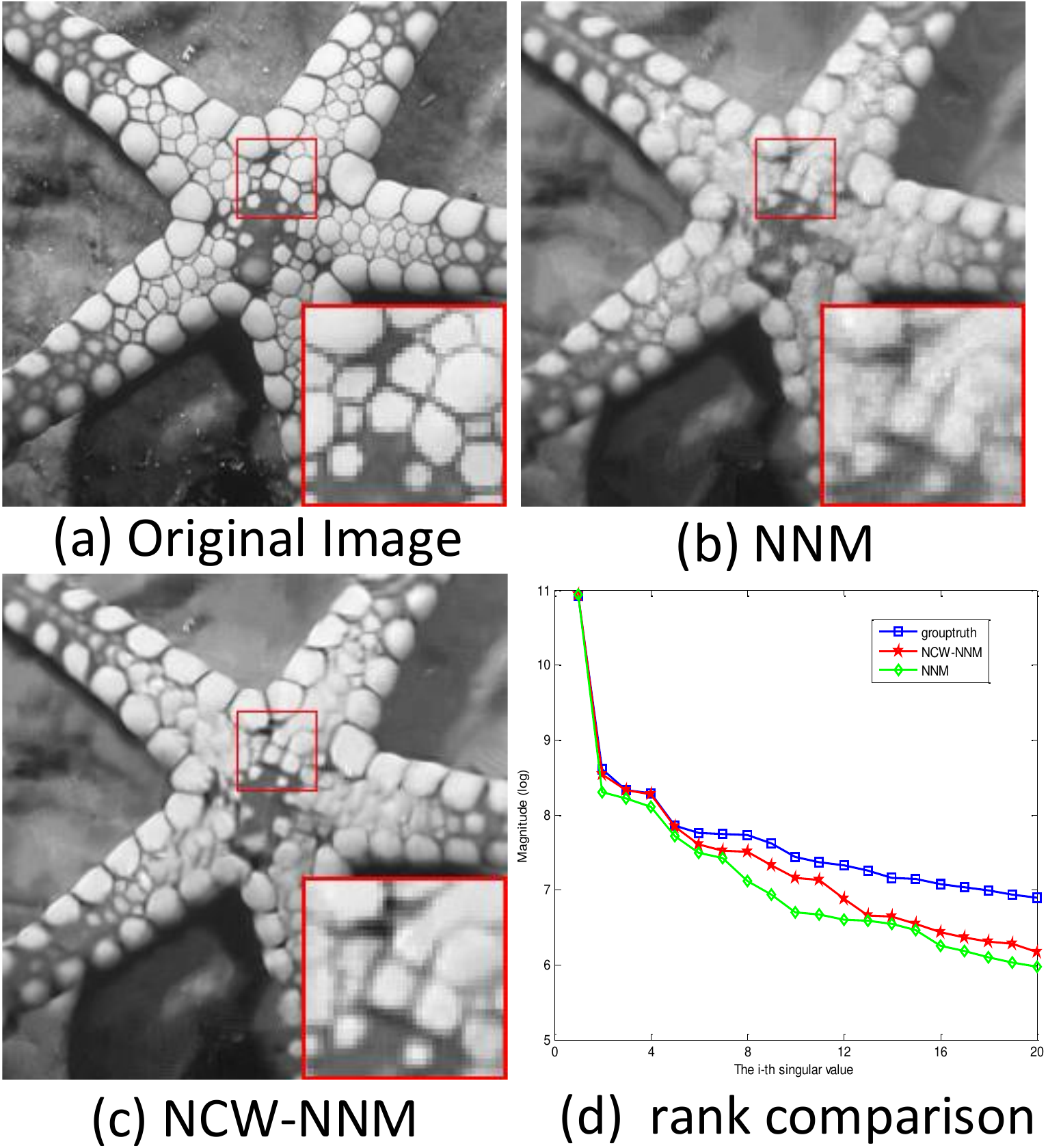}}
\end{minipage}
\caption{Visual comparison of \emph{Leaves} by image CS recovery with 0.1$N$ measurements. (a) Original image;  (b) NNM  (PSNR = 24.21dB, FSIM = 0.8484); (c) NCW-NNM (PSNR = \textbf{25.49dB}, FSIM = \textbf{0.8755}); (d) Comparison the rank of NNM and NCW-NNM methods.}
\label{fig:12}
\end{figure*}

In this subsection, to demonstrate the proposed NCW-NNM can improve the accuracy of the rank approximation effectively, we compare it with traditional nuclear norm minimization (NNM) method. In terms of the quantitative metrics, one can observe that the PSNR and FSIM results of the NNM  and the proposed NCW-NNM methods for image deblurring, image inpainting and image CS recovery are shown in Table~\ref{lab:2}, Table~\ref{lab:3} and Table~\ref{lab:4}, respectively. It can be seen that NCW-NNM consistently outperforms the NNM  method based on all IR tasks. The visual comparison results of image deblurring, image inpainting and image CS recovery are shown in Fig.~\ref{fig:10}, Fig.~\ref{fig:11} and Fig.~\ref{fig:12}, respectively. It can be seen that the proposed NCW-NNM produces much clear and better visual results than NNM method.

Moreover, Fig.~\ref{fig:10} (d), Fig.~\ref{fig:11} (d) and Fig.~\ref{fig:12} (d) show the singular values of these two rank approximation methods, respectively, where the data matrices are generated by selecting 60 similar image patches in accordance with the small red square exemplar patch. It can be seen that the singular values of the proposed NCW-NNM result is the best approximation to the grouptruth in comparison with NNM method. Therefore, the proposed NCW-NNM can enforce more accurate rank approximation results than NNM method. The main reason is that NNM method tends to over-shrink the rank components and treats each rank component equally, which cannot obtain the approximation of the matrix rank accurately. Different from NNM method, the proposed NNW-NNM assigns the different weight to each singular value and avoids the over-shrink phenomena. Accordingly, the proposed NCW-NNM can obtain more accurate rank approximation results than NNM method.
 \begin{figure*}[!htbp]
\begin{minipage}[b]{1\linewidth}
  \centerline{\includegraphics[width=12cm]{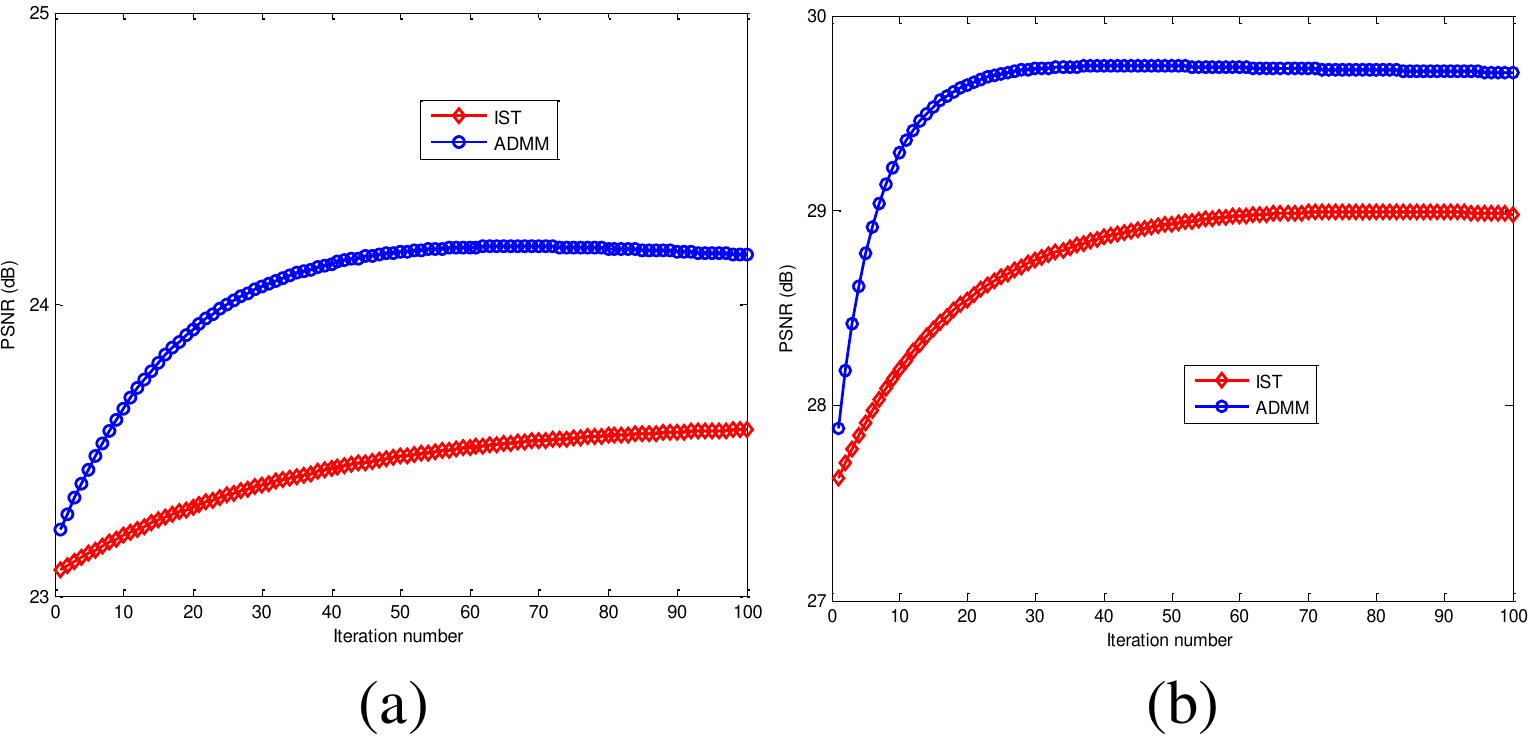}}
\end{minipage}
\caption{Comparison between ADMM and IST. (a) PSNR results achieved by ADMM and IST with ratio =0.2 for image $\emph{fingerprint}$. (b) PSNR results achieved by ADMM and IST with ratio =0.3 for image $\emph{pentagon}$.}
\label{fig:13}
\end{figure*}

\subsection{Comparsion Between ADMM and IST}

In this subsection, the classical optimization method: iterative shrinkage/theresholding (IST) \cite{68} is used to solve our proposed non-convex model for CS image reconstruction. We will make a comparison between ADMM and IST with ratio = 0.2 and ratio= 0.3 for two image $\emph{fingerprint}$ and $\emph{pentagon}$ as examples, respectively. Fig.~\ref{fig:13} shows the progression curves of the PSNR (dB) results achieved by ADMM and IST, respectively. It can be seen that ADMM algorithm is more fast efficient and effective to solve the proposed non-convex model than traditional IST algorithm.
\subsection{Convergence}
\begin{figure*}[!htbp]
\begin{minipage}[b]{1\linewidth}
  \centerline{\includegraphics[width=12cm]{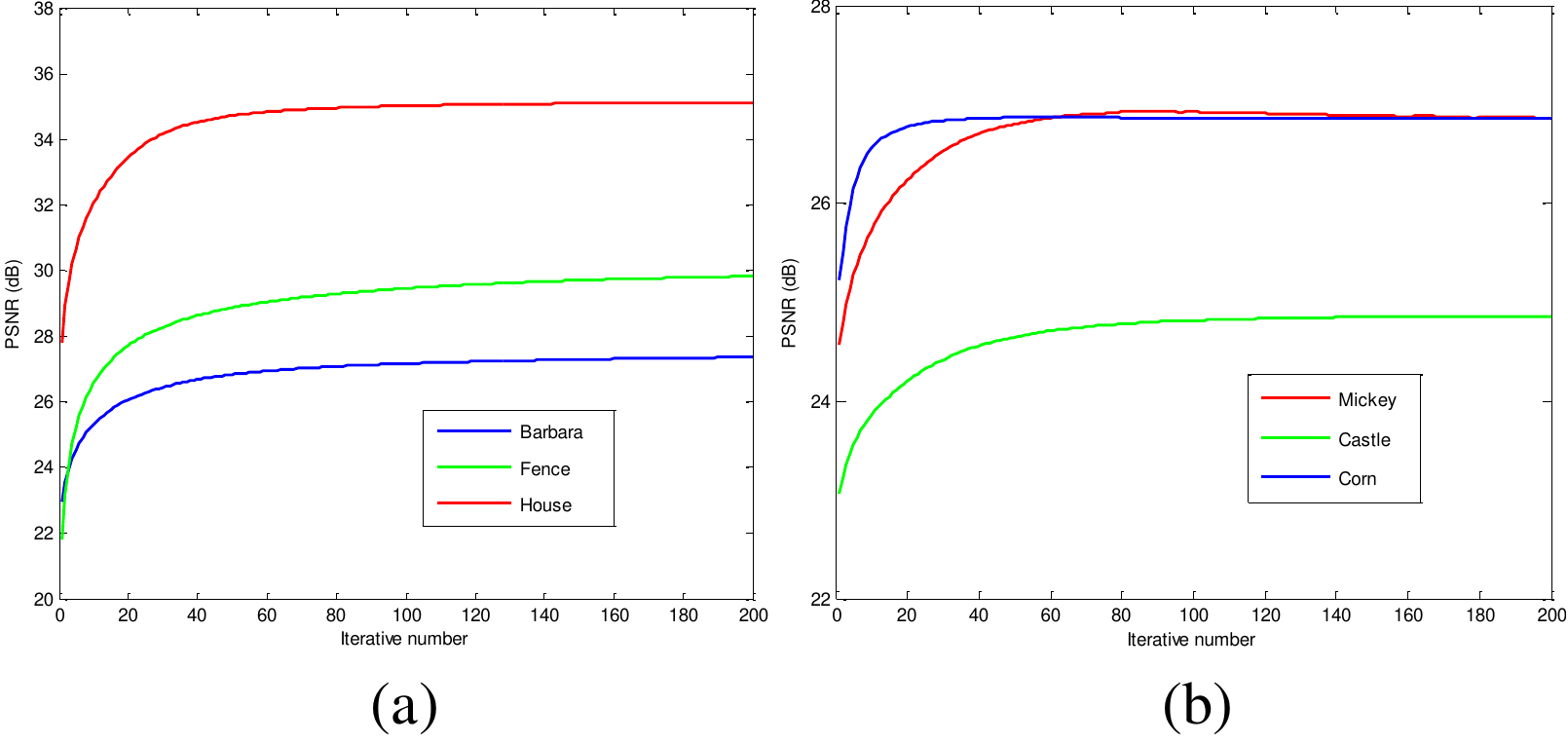}}
\end{minipage}
\caption{Convergence analysis of the proposed approach. (a) PSNR results versus iteration number for image deblurring with $9\times 9$ Uniform Kernel, $\delta=\sqrt{2}$. (b) PSNR results versus iteration number for image inpainting with 80\%  pixels missing.}
\label{fig:14}
\end{figure*}

Since the proposed model is non-convex, it is difficult to give its theoretical proof for global convergence. Here, we only provide empirical evidence to display the good convergence of the proposed NCW-NNM method. Fig.~\ref{fig:14} illustrates the convergent performance of the proposed NCW-NNM. It shows the curves of the PSNR values versus the iteration numbers for image deblurring with $9\times 9$ Uniform Kernel, $\delta=\sqrt{2}$ as well as image inpainting with 80\% pixels missing, respectively. One can observe that with the increase of the iteration numbers, the PSNR curves gradually increase and ultimately become flat and stable, and thus exhibiting the good stability of the proposed NCW-NNM method.
 \section{Conclusion}
 \label{5}
Image priors based on nonlocal self-similarity (NSS) and low-rank matrix approximation (LRMA) have achieved a great success in image restoration. However, since the singular values have clear meanings and should be treated differently, traditional nuclear norm minimization (NNM) regularized each of them equally, which often restricted its capability and flexility. To rectify the shortcoming of the nuclear norm, this paper proposed a new method for image restoration via non-convex weighted $\ell_p$ nuclear norm minimization (NCW-NNM). To make the optimization tractable, the alternative direction multiplier method (ADMM) framework was used to solve the proposed non-convex model. Experimental results on three image restoration applications, image deblurring, image inpainting and image CS recovery, have shown that the proposed method outperforms many current state-of-the-art methods both quantitatively and qualitatively.

\end{document}